\documentclass[10pt,twocolumn,letterpaper]{article}

\usepackage{iccv}
\usepackage{times}
\usepackage{epsfig}
\usepackage{graphicx}
\usepackage[ruled,lined]{algorithm2e}
\iccvfinalcopy
\def\iccvPaperID{80} 
\usepackage{microtype}
\usepackage{booktabs} % for professional tables
\usepackage{subcaption}
\usepackage{hyperref}

\usepackage{amsmath}
\usepackage{amssymb}
\usepackage{mathtools}
\usepackage{amsthm}

\usepackage[capitalize,noabbrev]{cleveref}
\usepackage{amsmath,amsfonts,bm}

\def\eqref#1{equation~\ref{#1}}
\def\1{\bm{1}}

\DeclareMathAlphabet{\mathsfit}{\encodingdefault}{\sfdefault}{m}{sl}
\SetMathAlphabet{\mathsfit}{bold}{\encodingdefault}{\sfdefault}{bx}{n}

\usepackage{hyperref}
\usepackage{url}
\usepackage{caption}
\usepackage[toc,page]{appendix}
\usepackage{graphicx} 
\usepackage{tcolorbox}
\tcbuselibrary{skins}
\usepackage{lipsum}
\usepackage{multirow}
\usepackage{tabularx}
\usepackage{booktabs}
\usepackage{diagbox}
\usepackage{wrapfig}
\ificcvfinal\fi
\begin{document}
\title{Dynamic Neural Network is All You Need: Understanding the Robustness of Dynamic Mechanisms in Neural Networks }
\author{Mirazul Haque, Wei Yang\\
University of Texas at Dallas, USA\\
{\tt\small mirazul.haque@utdallas.edu}
}
\maketitle
\begin{abstract}

%Deep Learning (DL) based solutions are being used to solve different problems faced by humans. Automatic Speech Recognition (ASR) is one of the DL-based technique that converts speech signals to natural language texts. ASR can be used in scenarios like helping visually impaired persons to send written documents. As the applications of ASR are real time based, it is important that the response time of the ASR systems is significantly low and also the system should be energy efficient. However, we find out that the response time of an ASR model is adaptive based on input. Hence, efficiency robustness of ASR models is needed to be evaluated. For that purpose, we propose \tool{}, a novel efficiency attack technique on ASR models. We find that efficiency of the ASR models are dependent on the length of output text, which is dependent on the input speech. Based on the finding, in \tool{}, we propose two approaches to modify the input such that the length of the outputs sequence is increased. We evaluate our approach on two ASR models and two popular datasets.

%such of the deep learning technique that has 

Deep Neural Networks (DNNs) have been used to solve different day-to-day problems. Recently, DNNs have been deployed in real-time systems, and lowering the energy consumption and response time has become the need of the hour. To address this scenario, researchers have proposed incorporating dynamic mechanism to static DNNs (SDNN) to create Dynamic Neural Networks (DyNNs) performing dynamic amounts of computation based on the input complexity. Although incorporating dynamic mechanism into SDNNs would be preferable in real-time systems, it also becomes important to evaluate how the introduction of dynamic mechanism
impacts the robustness of the models.
However, there has not been a significant number of works focusing on the robustness trade-off between SDNNs and DyNNs.
 To address this issue, we propose to investigate the robustness of dynamic mechanism in DyNNs and how dynamic mechanism design impacts the robustness of DyNNs.
  For that purpose, we evaluate three research questions. These evaluations are performed on three models and two datasets. Through the studies, we find that attack transferability from DyNNs to SDNNs is higher than attack transferability from SDNNs to DyNNs.
  Also, we find that DyNNs can be used to generate adversarial samples more efficiently than SDNNs. Then, through research studies, we provide insight into the design choices that can increase robustness of DyNNs against the attack generated using static model.
  Finally, we propose a novel attack to understand the additional attack surface introduced by the dynamic mechanism and provide design choices to improve robustness against the attack.
  
  %decrease the effectiveness of the DyNNs and can be used to evaluate design choices in DyNN.

\end{abstract}

\section{Introduction}

Deep Neural Networks (DNNs) are used in multiple applications such as computer vision and natural language processing. After the rapid growth of IoT and embedded devices, many real-time systems use DNNs in their applications. As the real-time systems require faster response time and low energy consumption, researchers have proposed to incorporate energy-saving dynamic mechanism~\cite{wang2018skipnet, kaya2019shallow, wu2018blockdrop} to popular static DNN (SDNN) models like ResNet~\cite{He2015}, VGG~\cite{simonyan2014very}, MobileNet~\cite{mobilenets} etc. Early-exit is one of the dynamic mechanism techniques where multiple exits are included in SDNNs (creating multiple sub-networks), and SDNNs can terminate the operation early if a certain sub-network is confident about the prediction. These types of DNNs are named as early-exit Dynamic Neural Networks (DyNNs).
Although the transition from SDNNs to DyNNs is preferred in real time systems because of increased efficiency, whether the use of dynamic mechanism will impact the robustness of the systems is still unknown. Studying the impact of the dynamic mechanisms on the robustness is important for developers or users to understand the trade-offs between DyNN and SDNN.

\begin{figure}
%\begin{wrapfigure}{r}{0.5\textwidth}%
	%\centering
	\begin{center}
	    \includegraphics[width=0.35\textwidth]{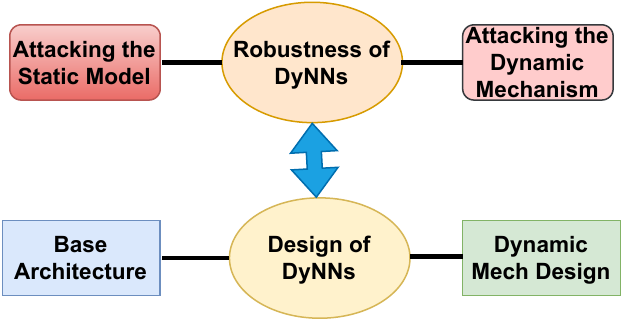}
	\end{center}
	
    \caption{The overview of the objective of this work.}
	\label{fig:overview_paper}
\end{figure}

 %In this work, we propose to investigate four different aspects of including dynamic mechanism through four research questions. These four aspects are: \textit{Transferability, Impact on Efficiency, Early-exits Design} and, \textit{Added Attack Surface}.

 In this work, we propose to investigate how robustness is impacted through the addition of dynamic mechanism to the static model and how different dynamic mechanism designs would impact the robustness in dynamic mechanism (Figure \ref{fig:overview_paper}). We propose to investigate two aforementioned topics through three research questions. These three RQs are based on: \textit{Robustness of Dynamic Mechanism, Robust Design for the Static Model Attack, and Robust Design for the Dynamic Mechanism Attack.}
 
 %mainly two aspects of including dynamic mechanism to the static model: robustness and the DyNN design. We specifically design three research questions two understand theses two aspects. These three RQs are based on:

 %\textit{Transferability, Impact on Efficiency, Early-exits Design} and, \textit{Added Attack Surface}.
 
 %four different aspects of including dynamic mechanism through four research questions. These four aspects are: \textit{Transferability, Impact on Efficiency, Early-exits Design} and, \textit{Added Attack Surface}.

 \noindent\textbf{Robustness of Dynamic Mechanism.} We focus on the investigation of robustness of the dynamic mechanism based on two aspects: transferability and the efficiency robustness.

 First, we investigate the adversarial attack transferability between SDNNs and DyNNs to evaluate the robustness of the dynamic mechanism in black-box scenarios. In the black-box scenarios, adversaries normally assume the target models are always static. However, the target models can be dynamic also. Hence, it is important to find out if a surrogate SDNN model is used to attack a target DyNN model or vice-versa, then, to which extent the adversary can be successful. 

To address this issue, in this paper, we first conduct a comparative study on the adversarial attack transferability between SDNNs and DyNNs (Section~\ref{sec:robustness_blackbox}). Our study results suggest that adversarial transferability from DyNNs to SDNNs is better and surprisingly, using DyNNs as surrogate models for attack seems to be a more efficient and more effective way to generate adversarial samples. The adversaries are able to generate more adversarial samples in the same amount of time compared to using SDNNs as the surrogate model, and the generated adversarial samples often can also attack SDNNs.

%After understanding that the introduction of the dynamic mechanisms generally will not reduce the robustness of the systems

The next robustness study is to understand how robust the efficiency of the dynamic mechanism is against the adversarial samples generated on SDNNs. This study tries to evaluate
whether the original purpose of DyNNs (\textit{i.e.,} saving inference time) will be impacted by the adversarial samples (Section~\ref{sec:energy_robustness}) generated through SDNNs.  Our study results suggest that the adversarial samples generated by existing white-box attacks and black-box attacks do not increase the inference time significantly, making the dynamic mechanisms efficiently robust against attacks on SDNNs.

%After understanding the robustness of DyNNs in general

\noindent\textbf{Robust Design for the Static Model Attack} Here, we perform a detailed analysis of which design choices in the dynamic mechanisms or DyNN architectures (specifically the position of early exits) may impact the robustness of DyNNs (Section~\ref{sec:analysis}). We consider two attack scenarios in this study: first, the output layer label of an SDNN is modified by a white-box adversarial example, and we study the impact of the example on corresponding DyNN's early-exit layers; second, in a black-box scenario, the output of SDNN is modified by a sample, and the sample is fed to separate model's DyNN. We have made multiple findings based on the empirical results, for example, putting the first exit earlier in the model architecture can help to improve the robustness of DyNNs.

\noindent\textbf{Robust Design for the Dynamic Mechanism Attack.} 
Last but not least, we evaluate the design choices of dynamic mechanism against a novel attack on dynamic mechanism.
First, we design an adversarial attack approach to understand the extra attack surface introduced by the dynamic mechanisms in neural network (Section~\ref{sec:attack}). In this attack, the synthesized adversarial examples will not change the prediction of the final output layer's label, but will change the prediction of all the early exits. Based on the evaluation results, we find that the dynamic mechanism is more vulnerable in scenarios where dependency among DyNN layers is lesser and when the exits are sparse w.r.t the layers.

\noindent\textbf{Contribution.} Our work provides insights about the robustness of the dynamic mechanism in DyNNs and also evaluates different design choices w.r.t robustness. The findings of this work would motivate developers/researchers to increase the usage of DyNNs and find more robust dynamic mechanisms.

% Our paper makes the following contributions:

% %\begin{itemize}
% %    \item \textbf{Empirical Novelty} Four research studies to understand the robustness of dynamic mechanisms in DyNNs. Based on findings of our studies, we make suggestions to design a more robust dynamic mechanism for DyNNs. 
%%     \item \textbf{Technical Novelty} A novel attack investigating the robustness of dynamic mechanisms in DyNNs. We consider a natural constraint (\textit{i.e.,} $\epsilon$) of perturbation. 
% \end{itemize}

% especially needs to be evaluated against robustness of SDNNs systematically to ensure safety. 

% However, there has not been any work focusing upon comparing the robustness of two type of models. If the early exit DyNNs networks are less robust than SDNNs, then it would impact the safety of the real time systems. In this work, we focus on the comparison between the robustness of DyNNs and SDNNs through three research questions. 

% First research question focuses on the transferability between adversarial samples generated of DyNNs and SDNNs.
\section{Related Works and Background}

\textbf{Dynamic Neural Networks.} The main objective of DyNNs is to decrease the energy consumption of the model inference for inputs that can be classified with fewer computational resources. DyNNs can be classified into \textit{Conditional-skipping DyNNs} and \textit{Early-exit DyNNs}. Early-exit DyNNs use multiple exits (sub-networks) within a single model and because of the model's working mechanism, the model is more suited for resource constrained devices.  
If, at any exit, the confidence score of the predicted label exceeds user defined threshold, inference is stopped.  The resource-constrained devices usually deploy a lightweight sub-network of early exit network locally and resort to a server for further computations if needed~\cite{teerapittayanon2017distributed} . \cite{graves2016adaptive}, \cite{figurnov2017spatially}, \cite{teerapittayanon2016branchynet}, \cite{kaya2019shallow} have proposed Early-termination DyNNs. Specifically \cite{kaya2019shallow} and \cite{zhou2020bert} propose early exit networks based on popular SDNNs. \cite{zhou2020bert} also show that white-box robustness of the DyNNs is better than SDNNs. In addition to that multiple works~\cite{teerapittayanon2017distributed,scardapane2020should} provide practical usability of DyNNs.
%Detailed explanation of early-exit networks can be found at Appendix (\ref{sec:early}).
\begin{figure}
%\begin{wrapfigure}{r}{0.5\textwidth}%
	%\centering
	\begin{center}
	    \includegraphics[width=0.45\textwidth]{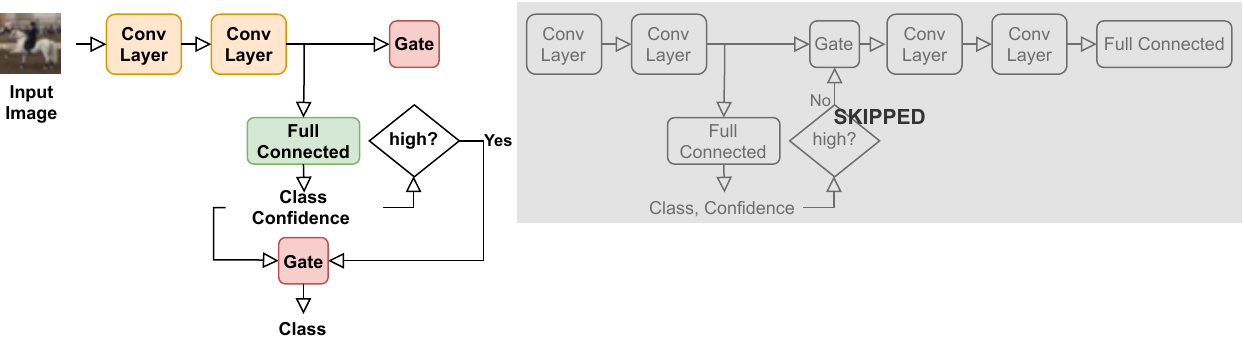}
	\end{center}
	
    \caption{Working mechanism of Early-exit DyNN}
	\label{fig:example2}
\end{figure}

 Figure \ref{fig:example2} shows the working mechanism of Early-exit DyNNs.
 For example, an Early-exit DyNN has N parts and each part has an exit. $x$ is the input, $f^i_{out}$ represents prediction after the $i^{th}$ part (generated by specific computation unit), $f_{out}$ represents prediction of the Neural Network, $C_i$ represents confidence score after $i^{th}$ part, $Hid^{In}_i$ represents input of $i^{th}$ part, $Hid^{Out}_i$ represents output of $i^{th}$ part, and $\tau_i$ is the predefined threshold to exit the network after $i^{th}$ part. The working mechanism of the Early-exit DyNN can be represented as, $f_{out}(x)=f^i_{out}(x), \text{if}\ C_i(x) \ge \tau_i$.

\begin{comment}
\begin{equation} \label{eq:new3}
    \begin{cases}
      f_{out}(x)=f^i_{out}(x), \text{if}\ C_i(x) \ge \tau_i \\
      Hid^{In}_{i+1}(x)=Hid^{Out}_i(x),\text{otherwise}
    \end{cases}
 \end{equation}
\end{comment}

\textbf{Adversarial Attacks.} Adversarial Examples are the input that can change the prediction of the DNN model when those are fed to the model.
\cite{goodfellow2014explaining} propose Fast Gradient Sign Method (FGSM) that uses single-step first order entropy loss to generate adversarial inputs. This attack is modified by \cite{madry2017towards} to add initial noise to the benign sample. 
This attack is referred as projected gradient descent (PGD). 
Other than that, \cite{dong2018boosting,carlini2017towards,croce2020reliable,lin2019nestero} have proposed white-box attack methods, while \cite{liu2016delving,andriushchenko2020square,ilyas2018black,yang2022testaug} have proposed black-box attack methods. Recently, attacks~\cite{MirazILFO,haque2022ereba,haque2021nodeattack,haque2023slothspeech, haque2022corrgan, chen2022deepperform,chen2022nicgslowdown,chen2022nmtsloth,chen2023dark} have been proposed to decrease the efficiency of the dynamic neural networks.
%The DyNNs use specific computing units within DNN, and intermediate outputs generated by those computing units are used to disable certain parts of Neural Networks, decreasing the computational resources consumption. 
\section{How robust are DyNNs against adversarial examples?}
\label{sec:robustness_blackbox}

In this research question, we investigate the robustness of dynamic mechanism in DyNNs w.r.t adversarial examples generated on SDNNs. We further divide this question into two sub-parts: (i) attack transferability between SDNNs and DyNNs and (ii)  efficiency robustness of DyNNs against adversarial examples generated on SDNNs. Through these RQs, we get an insight about how the existing adversarial attacks on SDNNs would impact the accuracy and efficiency of DyNNs. Also, we find out how adversarial examples generated on DyNNs impact the SDNNs w.r.t accuracy. As the SDNN models are static, adversarial examples would not have any impact on the efficiency of SDNNs.

\subsection{Is adversarial example transferability from DyNN to SDNN  lower than adversarial example transferability from SDNN to DyNN?}
%What is the traditional accuracy attack transferability between AdNNs and SDNNs? }

%\usepackage{subfig}

%Accuracy based adversarial attacks have been used to evaluate the robustness of a DNN. Adversarial samples are generated by adding minimal perturbation to a benign input such that the prediction of the DNN becomes wrong for that particular input. 

In this research question, we investigate the ``transferability'' of adversarial inputs generated based on SDNN and DyNN, respectively, \textit{i.e.,} whether adversarial examples generated based on SDNNs are adversarial to DyNNs and vice versa. Transferability is an important metric for evaluating the feasibility of black-box attack. 
To evaluate the transferability, one of the popular way~\cite{papernot2017practical,liu2016delving} is creating a similar model (\textit{i.e.,} surrogate model) as the target model.
In a black-box attack, normally, adversaries assume the underlying model to be SDNN, so for a deployed DyNN, the adversaries may likely use an SDNN as surrogate model. Hence, this research question (RQ) is important to evaluate the robustness of DyNNs. 

%If attack transferability is higher in any particular direction (SDNN to DyNN or DyNN to SDNN), then the robustness of one specific type of model is regarded as lower.

%The intuition behind this RQ is: \textit{if attack transferability is higher in any particular direction (SDNN to DyNN or DyNN to SDNN), then the robustness of one specific type of model is lower.} If there is a deployed model attacker wants to attack without knowing the type of the model (SDNN or DyNN), our goal is to find out  which type of DNN is more vulnerable against black-box adversarial attack.

%0.33\textwidth
\begin{figure*}[t]
    \centering
    \begin{subfigure}[b]{0.23\textwidth}
        \includegraphics[width=\textwidth]{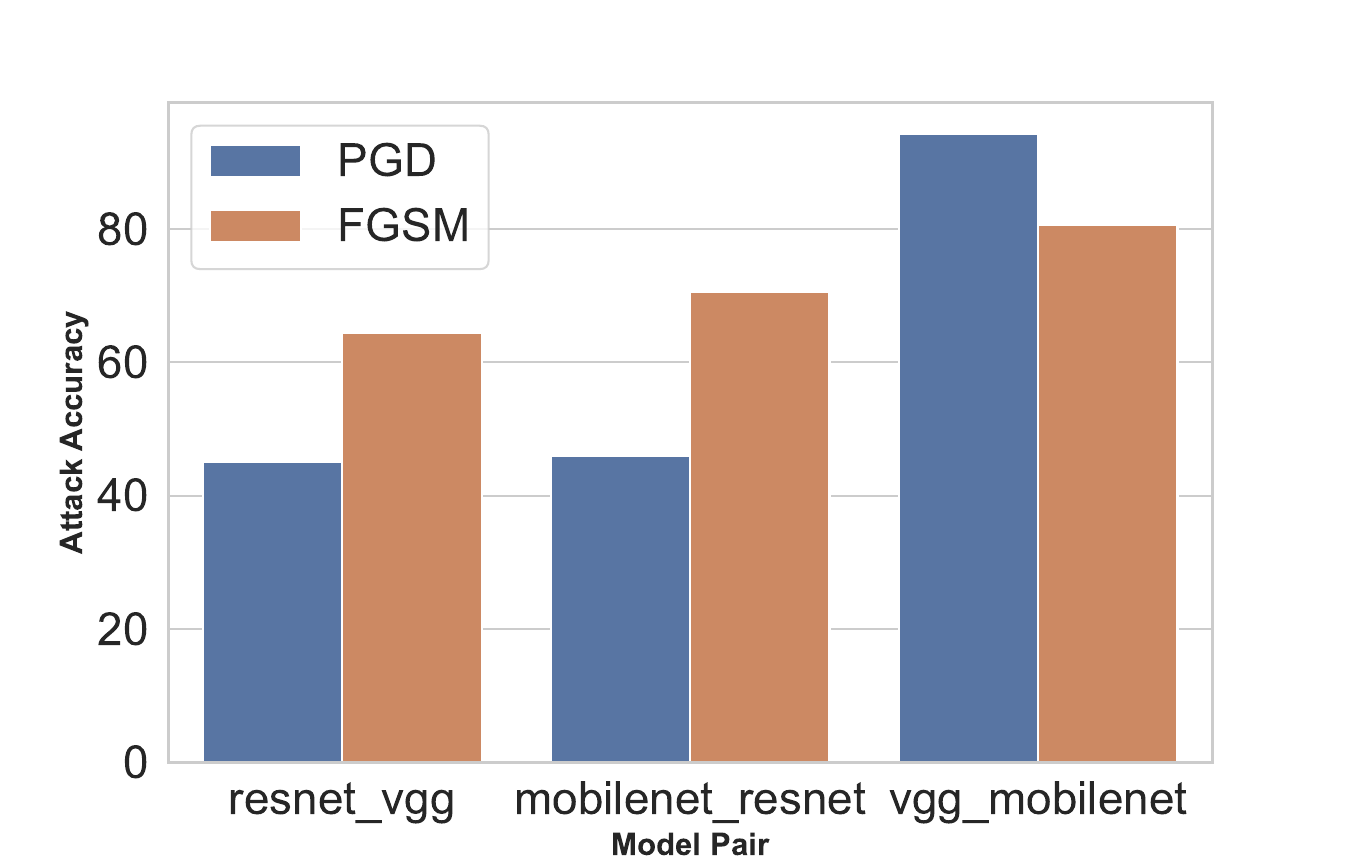}
        \caption{Target DyNN (CIFAR-10)}
    \end{subfigure}
    \begin{subfigure}[b]{0.23\textwidth}
        \includegraphics[width=\textwidth]{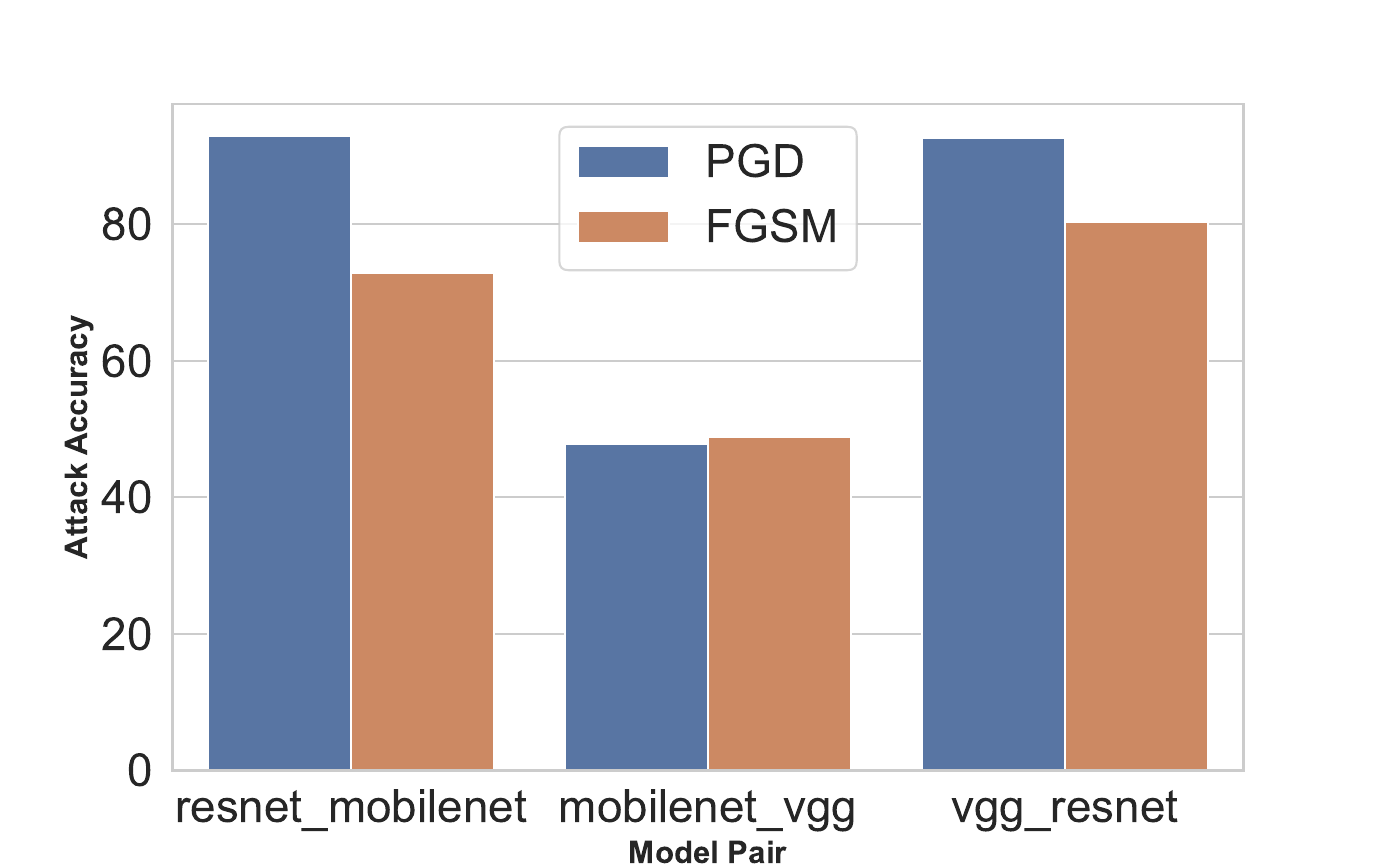}
        \caption{Target SDNN (CIFAR-10)}
    \end{subfigure}
    \begin{subfigure}[b]{0.23\textwidth}
        \includegraphics[width=\textwidth]{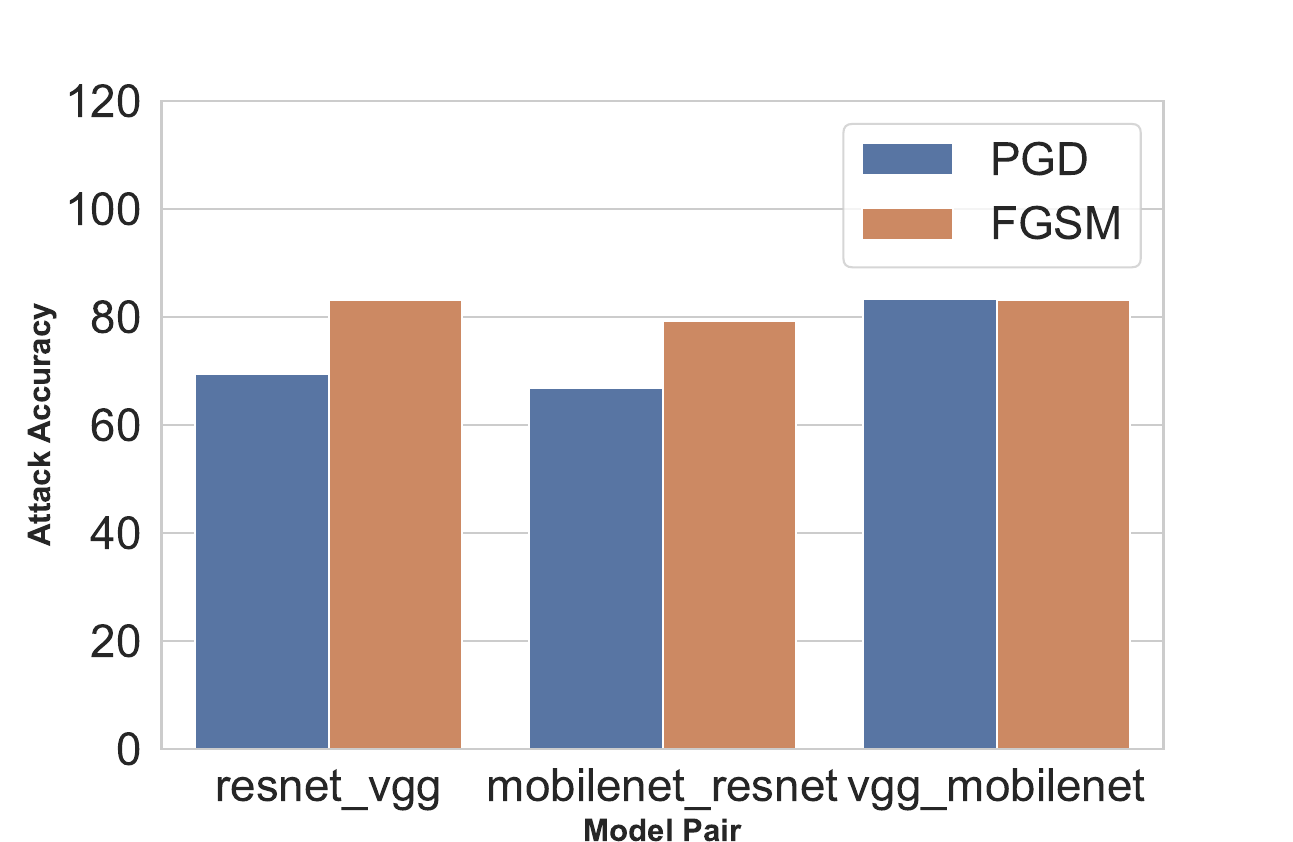}
        \caption{Target DyNN (CIFAR-100)}
    \end{subfigure}
    \begin{subfigure}[b]{0.23\textwidth}
        \includegraphics[width=\textwidth]{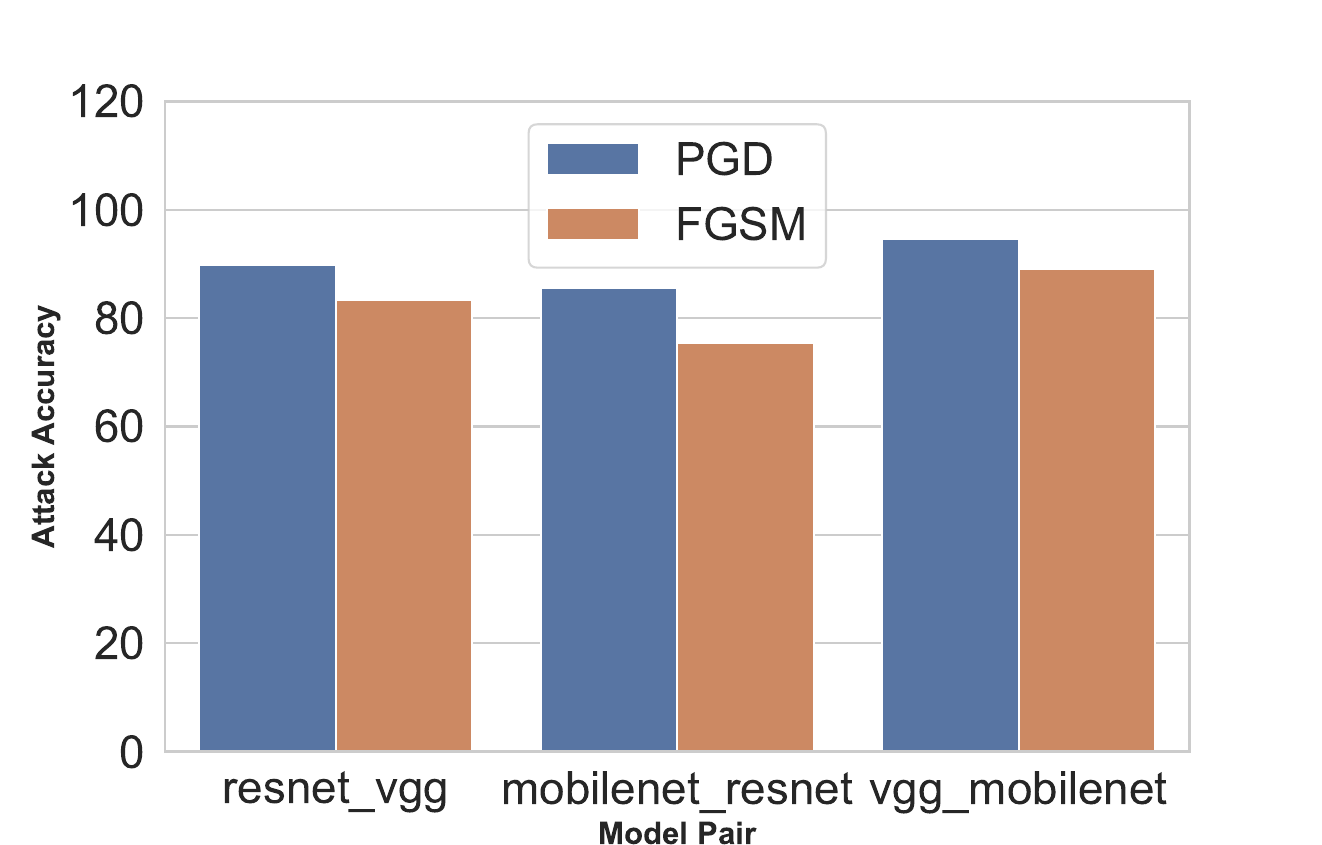}
        \caption{Target SDNN (CIFAR-100)}
    \end{subfigure}
    \caption{Transferable Attack Success Rate for CIFAR-10 and CIFAR-100 data}
    \label{fig:dicussion}
\end{figure*}

\subsubsection{Experimental Setup.} 

\textbf{Dataset and Models.} We use CIFAR-10 and CIFAR-100 \cite{krizhevsky2009learning} datasets for evaluation. For SDNNs, we use VGG-16 \cite{simonyan2014very}, ResNet56 \cite{He2015}, and MobileNet \cite{mobilenets} model. As DyNNs, we use the early exit version of these models~\cite{kaya2019shallow}. In all other RQs, we keep the models and dataset setup same.

\textbf{Black-box Attack.} For the attack scenario, we use surrogate model~\cite{papernot2017practical, liu2016delving} based black-box attack scenario. Here, we feed a set of inputs to the target model and collect the output labels. These inputs are generated using 50\% of the held-out validation data, and naturally corrupted versions of those validation data. As the number of partial held-out validation data is not significant, adding corrupted inputs would help to increase training data size for the surrogate model. For natural corruption~\cite{hendrycks2019benchmarking}, we use gaussian noise and brightness. For each type of corruption, we have five intensity levels. For example, if the number of held out data is 5000, for each corruption, we generate 25000 additional data. Once the input-output pairs are collected, a surrogate model is trained based on those pairs. For a target model, we use both SDNNs and DyNNs. If the target model is SDNN, then an DyNN is trained as surrogate model and vice-versa.

To make the surrogate-target pairs, we use different types of DNN architectures. For example, if the 
the target model is DyNN VGG, we choose ResNet56 SDNN as the surrogate model. This assumption is valid because the attacker doesn't have information about the target model architecture, hence the possibility of choosing the same architecture as the surrogate model is less. We define two terms to represent two different types of transferability based on different types of surrogate model and target model: \textbf{D2S} transferability and \textbf{S2D} transferability. D2S transferability evaluates DyNN to SDNN attack transferability, where S2D transferability evaluates SDNN to DyNN attack transferability.
We have chosen following pairs to evaluate S2D transferability: (SDNN ResNet56 (surrogate), DyNN VGG (target)), (SDNN MobileNet (surrogate), DyNN MobileNet (target)), (SDNN MobileNet (surrogate),DyNN ResNet56 (target)). Similarly, for D2S transferability, earlier mentioned surrogate models become the target model and earlier mentioned target models become the surrogate model. 

%We have chosen following pairs for DyNNs and SDNNs : DyNN VGG (target): SDNN ResNet56 (surrogate), DyNN ResNet56 (target): SDNN MobileNet (surrogate), and DyNN MobileNet (target): SDNN VGG (surrogate). Similarly, for SDNNs as the target, earlier mentioned surrogate models become the target model and earlier mentioned target models become the surrogate model. 

\textbf{Algorithms.} We use FGSM \cite{goodfellow2014explaining} and PGD \cite{madry2017towards} algorithms to attack the surrogate models.

\textbf{Metric.} We measure percentage of adversarial examples that can mis-classify the output w.r.t number of generated adversarial examples as the attack success rate.

\subsubsection{Evaluation Results}

Figure \ref{fig:dicussion} shows the effectiveness of black-box attacks on DyNNs and SDNNs. On average, it can be noticed that for target SDNN and surrogate DyNN, the attack success rate is higher than the success rate of target AdNN and surrogate SDNN. One of the reasons for this behavior is the lower variance of the DyNNs. DyNNs use lower number of parameters, hence the feature space for adversarial samples of DyNNs is smaller than the feature space for adversarial samples of SDNNs~\cite{schonherr2018adversarial}. Also, we find that for the target DyNN, FGSM attack performs better than PGD attack. If only dataset-specific results are considered, then for CIFAR-100 the attack success rate is higher than for CIFAR-10.

Also, as the DyNN-generated adversarial inputs can attack SDNNs, then it can be time efficient to create adversarial inputs using DyNN. Through Figure \ref{fig:rq1_2} (Appendix), we can find the probability density plots of different exit numbers of DyNN that have been used to generate adversarial examples. Lower exit number suggests that lesser number of computations has been used to generate adversarial examples. It can be noticed that for PGD attack, more than 70\% of adversarial examples are generated from exit 0, 1 and 2 (first - third exit). Although for FGSM attack, in a few scenarios (CIFAR-10 VGG, CIFAR-100 MobileNet, and CIFAR-100 ResNet), more than 50\% of adversarial examples are generated through later exits (higher computation required).

We also perform some additional experiments to strengthen our claim. In appendix, Figure 13 shows the experimental results on MI-FGSM attack, which is also a highly transferable attack. Also Figure 17 shows the results on Tiny-ImageNet dataset, which has a higher input size than CIFAR datasets. These two results show that our claim also holds for addditional attacks and datasets. Also Figure 14,15 and 16 in appendix shows adversarial examples generated through SDNNs and DyNNs. We find that both type of adversarial examples are similar w.r.t quality.

\begin{center}
\begin{tcolorbox}[colback=gray!10,%gray background
                  colframe=black,% black frame colour
                  width=8cm,% Use 8cm total width,
                  arc=1mm, auto outer arc,
                  boxrule=0.9pt,
                 ]
 \textbf{Finding 1}:  \textit{The \textbf{D2S} transferability is higher than the \textbf{S2D} transferability.}

 \textbf{Finding 2}:  \textit{Using DyNNs as surrogate models is more efficient and more effective way to generate adversarial examples than using SDNNs.}
\end{tcolorbox}
\end{center}

\subsection{Does adversarial examples impact efficiency robustness in DyNNs?}
\label{sec:energy_robustness}

In this section, we investigate if the dynamic mechanism in DyNNs is robust w.r.t efficiency against the adversarial examples generated on static mdoel.
Through the evaluation, we ensure whether the original purpose of including dynamic mechanism in DyNNs (\textit{i.e.,} saving inference time) will be impacted by the adversarial samples. Specifically, we study whether the adversarial inputs exit earlier or later in a DyNN compared to the original inputs. The main objective of this investigation is to find out whether the adversarial samples generated on SDNN can have an impact on the amount of computation of the DyNN. For this purpose, we conduct both white-box and black-box attacks to find out the impact of adversarial samples on the amount of computation.

Here, the white-box attack scenario can also be considered a practical scenario. There have been studies~\cite{chen2022learning, DnD} that focus on reverse engineering of SDNN models from binary code of on-device models, but no techniques have been proposed to reverse engineer the dynamic mechanisms in the models. Hence, adversaries are more likely to get SDNN models instead of their dynamic counterparts. So it is important to find out how adversarial samples affect the efficiency of the DyNN for both white-box and black-box scenarios.

\subsubsection{Experimental Setup.} 
%\textbf{Dataset and Models.} We use same datasets and models as previous RQ.

\textbf{Attack.} We use PGD and FGSM for both white-box and black-box attacks. For black-box setup, we use the same setup as previous RQ. For black-box scenario, we use DyNNs as target model and SDNNs as surrogate model. In a white-box setting, we attack the SDNN and evaluate on the performance of corresponding DyNN.

\textbf{Metric.} We use the difference between the exit number selected by adversarial input and the exit number selected by benign input. If the difference is positive, then the latency of the adversarial sample is increased w.r.t benign input.

\subsubsection{Evaluation Results.}

Figure \ref{fig:rq2_1} and Figure \ref{fig:rq2_1_appnd} (in Appendix) show the impact of adversarial examples generated on SDNN on changing in exit number in DyNN in a white-box setting. It can be observed that for the majority of the scenarios, accuracy-based adversarial samples do not increase the computation significantly in the DyNN. On average, FGSM-generated examples increase more computation in DyNNs than PGD-generated examples. For CIFAR-10 data, for MobileNet and VGG-16 DyNN models, 25\%-37\% of the FGSM generated examples could increase the number of exits by more than one.
Also, it can be noted that adversarial samples generated on CIFAR-100 data is more likely to increase computation than adversarial samples generated on CIFAR-10. Especially, more than 45\% of the FGSM samples generated on CIFAR-100 data can increase the number of exits by more than one.

Figure \ref{fig:rq2_2} and Figure \ref{fig:rq2_2_appnd} (in Appendix) show the impact of adversarial examples generated on SDNN on change in exits in DyNN in a black-box setting. It can be noted that, for CIFAR-10 dataset, black-box attack can generate more computation-increasing examples than in white-box attack. For ResNet and VGG model, more than 40\% PGD attack generated examples can increase the number of exits by more than one. For all three models, 35\% of the FGSM attack generated examples can increase the number of exits by more than one. FGSM attack generates more  inputs that induces low confidence in early-exit layers than PGD attack. For CIFAR-100 dataset, the increase of computation caused by adversarial attacks is higher than that of CIFAR-10. As CIFAR-100 data uses larger model, the robustness of the model is reduced. For CIFAR-100 dataset, more than 40\% examples generated through both the attack can increase the number of exits by more than one. However, increasing the number of exits by two or three exits does not decrease the efficiency in DyNNs significantly. 

%Hence, we find that accuracy-based adversarial samples do not increase the computation significantly in the DyNN.

\begin{center}
\begin{tcolorbox}[colback=gray!10,%gray background
                  colframe=black,% black frame colour
                  width=8cm,% Use 8cm total width,
                  arc=1mm, auto outer arc,
                  boxrule=0.9pt,
                 ]
 \textbf{Finding 3}:  \textit{Accuracy-based adversarial samples do not decrease the efficiency significantly in the DyNN.}
 
 \textbf{Finding 4}:  \textit{The adversarial examples, whose output confidences are significantly lower, can perform better in terms of decreasing the efficiency in DyNNs.}
 
 \textbf{Finding 5}:  \textit{Adversarial examples generated on a larger model (w.r.t model parameters) is more likely to decrease efficiency in DyNNs. }
\end{tcolorbox}
\end{center}

\begin{figure*}[t]
    \centering
    \begin{subfigure}[b]{0.28\textwidth}
        \includegraphics[width=\textwidth]{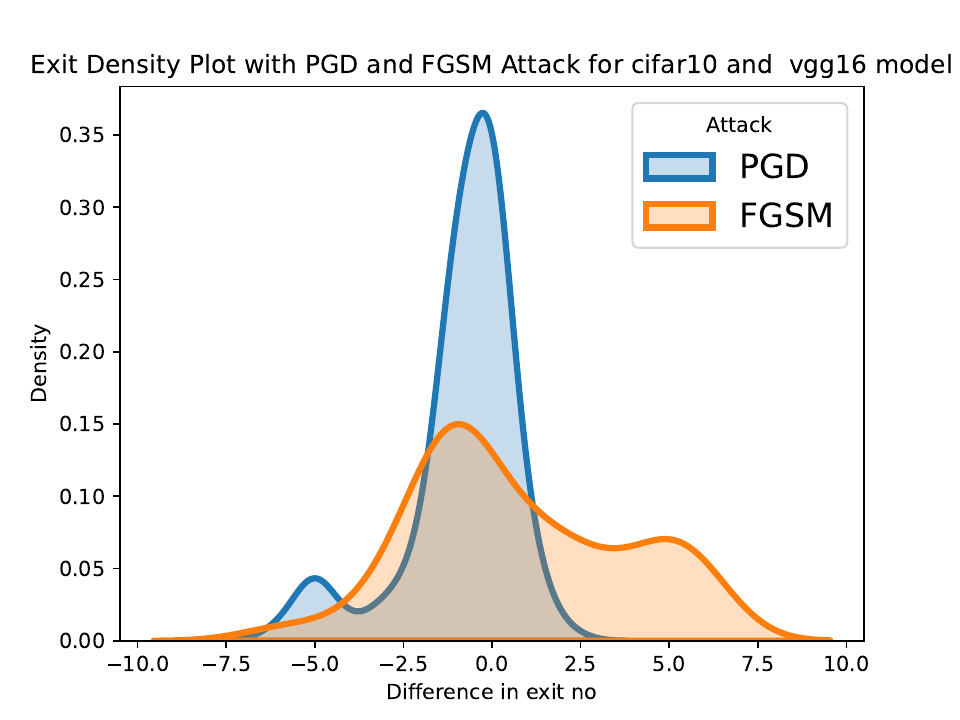}
        \caption{CIFAR-10 VGG}
    \end{subfigure}
    \begin{subfigure}[b]{0.29\textwidth}
        \includegraphics[width=\textwidth]{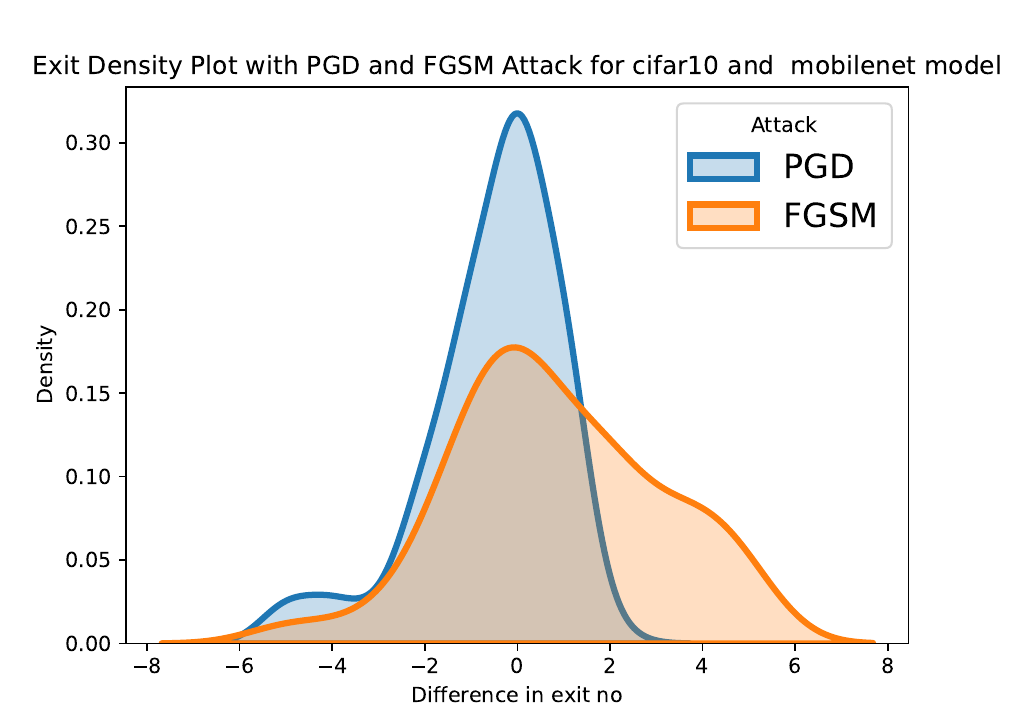}
        \caption{CIFAR-10 MobileNet}
    \end{subfigure}
    \begin{subfigure}[b]{0.28\textwidth}
        \includegraphics[width=\textwidth]{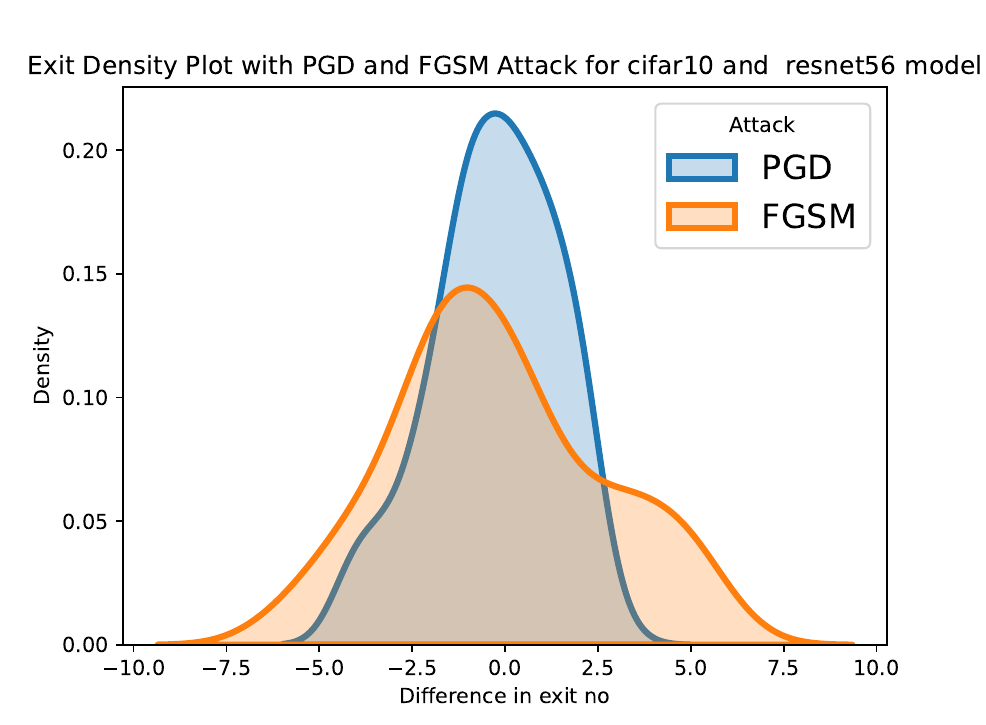}
        \caption{CIFAR-10 ResNet}
    \end{subfigure}
    \caption{Density plots of change in exit numbers because of PGD and FGSM attack (For CIFAR-10 data). The x axis represents the change in exit number while y axis represents the density.}
    \label{fig:rq2_1}
\end{figure*}

\begin{figure*}[t]
    \centering
    \begin{subfigure}[b]{0.28\textwidth}
        \includegraphics[width=\textwidth]{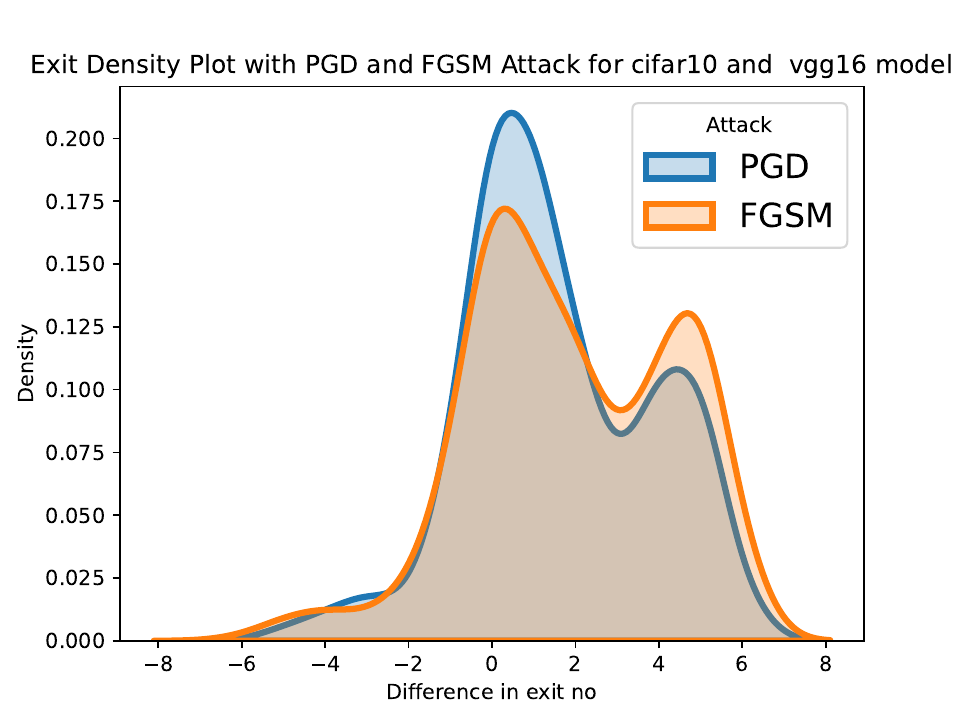}
        \caption{CIFAR-10 VGG}
    \end{subfigure}
    \begin{subfigure}[b]{0.29\textwidth}
        \includegraphics[width=\textwidth]{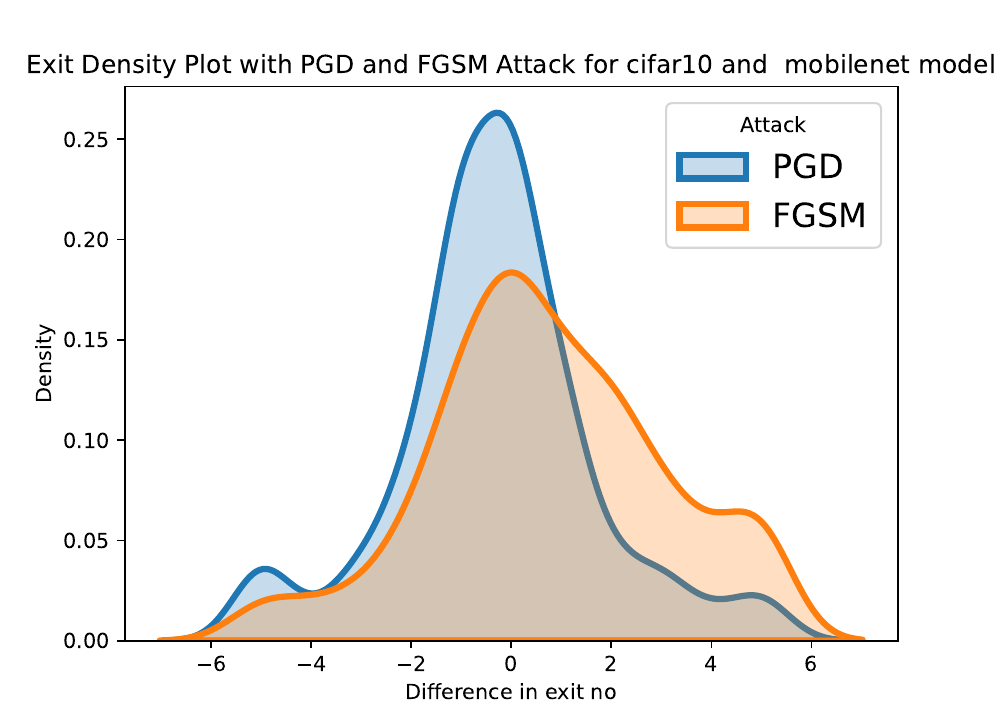}
        \caption{CIFAR-10 MobileNet}
    \end{subfigure}
    \begin{subfigure}[b]{0.28\textwidth}
        \includegraphics[width=\textwidth]{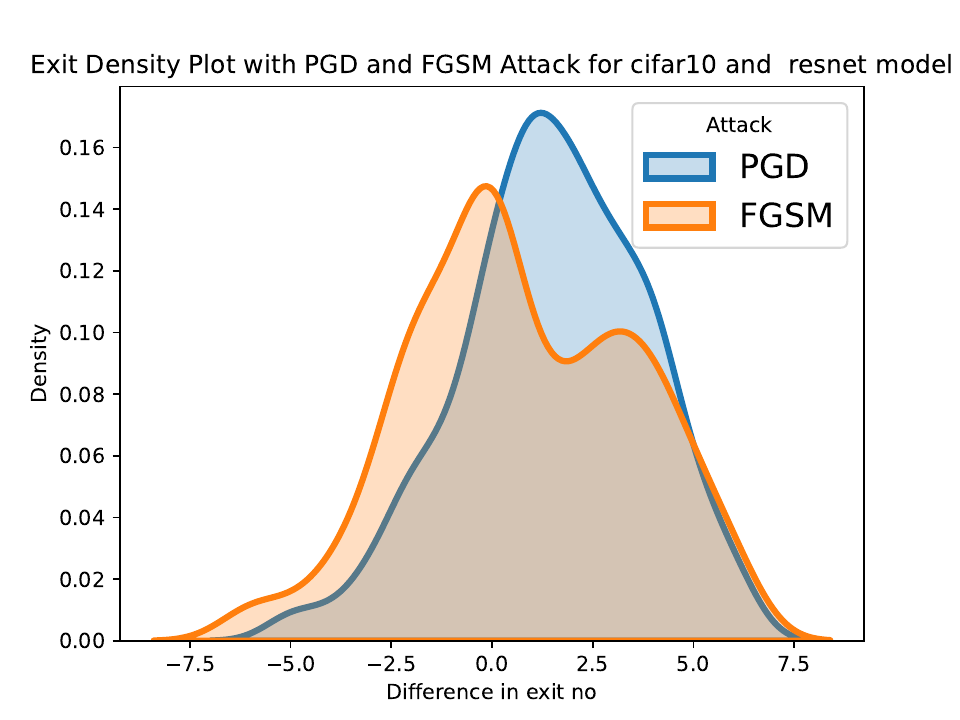}
        \caption{CIFAR-10 ResNet}
    \end{subfigure}
    
    \caption{Density plots of change in exit numbers because of PGD and FGSM \textbf{black-box} attack (For CIFAR-10). The x axis represents the change in exit number while y axis represents the density.}
    \label{fig:rq2_2}
\end{figure*}

\begin{comment}
\begin{subfigure}[b]{0.28\textwidth}
        \includegraphics[width=\textwidth]{figs/c100_vgg_bb.pdf}
        \caption{CIFAR-100 VGG}
    \end{subfigure}
    \begin{subfigure}[b]{0.29\textwidth}
        \includegraphics[width=\textwidth]{figs/c100_mnet_bb.pdf}
        \caption{CIFAR-100 MobileNet}
    \end{subfigure}
    \begin{subfigure}[b]{0.28\textwidth}
        \includegraphics[width=\textwidth]{figs/c100_rnet_bb.pdf}
        \caption{CIFAR-100 ResNet}
    \end{subfigure}
    
    \begin{subfigure}[b]{0.28\textwidth}
        \includegraphics[width=\textwidth]{figs/c100_vgg.pdf}
        \caption{CIFAR-100 VGG}
    \end{subfigure}
    \begin{subfigure}[b]{0.29\textwidth}
        \includegraphics[width=\textwidth]{figs/c100_mnet.pdf}
        \caption{CIFAR-100 MobileNet}
    \end{subfigure}
    \begin{subfigure}[b]{0.29\textwidth}
        \includegraphics[width=\textwidth]{figs/c100_rnet.pdf}
        \caption{CIFAR-100 ResNet}
    \end{subfigure}
\end{comment}

\section{What design of DyNNs may impact the robustness against attack generated on SDNNs?}
\label{sec:analysis}

%How adversarial examples impact the prediction at different exits

In this section, we evaluate which architecture design choices (position of early exits) in DyNNs  may impact the robustness of early layers against the adversarial inputs generated on SDNNs. In this RQ, we consider DyNNs as multi-exit networks, where each exit will provide an output. 
%Our main objective is to evaluate for which architecture design choices (position of early exits), the robustness of early layers is better. 
For evaluation, we assess if we attack the final exit (output of the static model), from which early-exit layer the label modification begins.  
 If the output label is not modified in any earlier layer, then for that type of model, the robustness is higher because the model can produce correct results at any layer. This RQ will provide us an insight into which type of design choice may improve the robustness of early layers.

%the robustness of earlier exits in an DyNN given an adversarial example generated for SDNN. In previous RQs, we only consider the exit in DyNN which provides the final output based on high confidence score. 

\subsection{Experimental Setup.} 
%\textbf{Dataset and Models.} We use same datasets and models as RQ1 and RQ2.

\textbf{Attack Setup.} We use same attack setup as previous research questions. However, while we attack the DyNN, we do not consider one specific exit layer. Instead of that, we consider each exit layer.

\textbf{Metric.} We use the early exit from which the label modification starts. For example, there are $N$ exits. First, $N$th exit's label is modified through adversarial sample. If till $K$th exit the original prediction was same, then we report $K+1$th exit in the experimentation.

\subsection{Evaluation Results.} 

Figure \ref{fig:rq3_1} and Figure \ref{fig:rq3_1_appnd} (in Appendix) show probability density plot on which exit the output label is changed using the white-box adversarial examples. It can be observed that for all the model-dataset pairs, for more than 77\% of the examples, the label is modified in the first exit. For CIFAR-10 data, only for MobileNet and ResNet models, the label is changed after the first exit for more than 20\% of the examples (using FGSM attack).

Figure \ref{fig:rq3_2} and Figure \ref{fig:rq3_2_append} (in Appendix) show probability density plot on which exit the output label is changed using the black-box adversarial examples. From results, we can see that the robustness of earlier exits is better against black-box attack than white-box attack. For CIFAR-10 data, more than 45\% of the both attack generated samples could not misclassify the first exit for VGG-16 model. For CIFAR-100 data (larger model), robustness of early exit layers is worse than that of CIFAR-10 data (smaller model). For CIFAR-100 data, for both attacks and three models, less than 30\% of the adversarial examples can not mis-classify the first exit.  

For black-box attack scenario, VGG-16 model's first layer on-average shows better robustness against black-box attack. In the DyNN, the first exit of the VGG model is placed only after second layer, while for others, more computations are performed before using the first exit. Hence, having the first exit in the early layers can increase the robustness of DyNN. Although lower number of parameters would be used to predict output, but with VGG we can notice that a significant number of inputs can be classified correctly through first exit.

\begin{center}
\begin{tcolorbox}[colback=gray!10,%gray background
                  colframe=black,% black frame colour
                  width=8cm,% Use 8cm total width,
                  arc=1mm, auto outer arc,
                  boxrule=0.9pt,
                 ]
 \textbf{Finding 6}:  \textit{We find that having the first exit in the earlier layers can increase the robustness of early exits of the DyNNs.}
 
 \textbf{Finding 7}:  \textit{Black-box attack success rate is lesser than white-box attack success rate against early-exit layers.}
 
 \textbf{Finding 8}:  \textit{For black-box scenario, early-exit layers are more robust against adversarial examples generated on a smaller surrogate model than adversarial examples generated on a larger surrogate model.}

\end{tcolorbox}
\end{center}

\begin{figure*}[t]
    \centering
    \begin{subfigure}[b]{0.28\textwidth}
        \includegraphics[width=\textwidth]{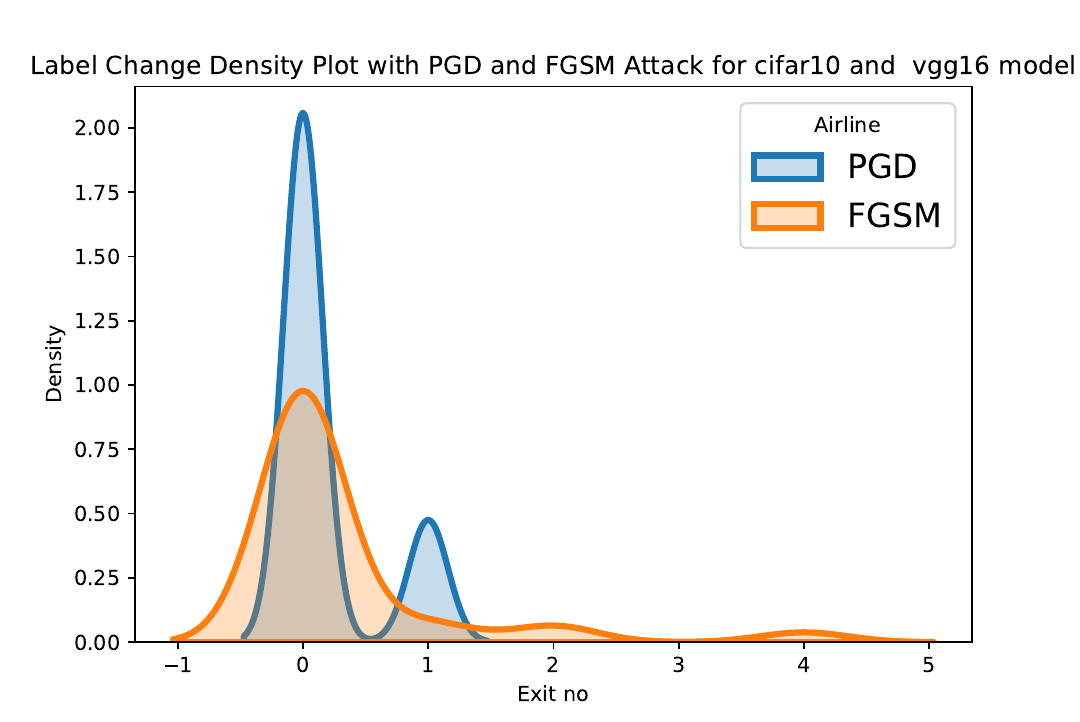}
        \caption{CIFAR-10 VGG}
    \end{subfigure}
    \begin{subfigure}[b]{0.29\textwidth}
        \includegraphics[width=\textwidth]{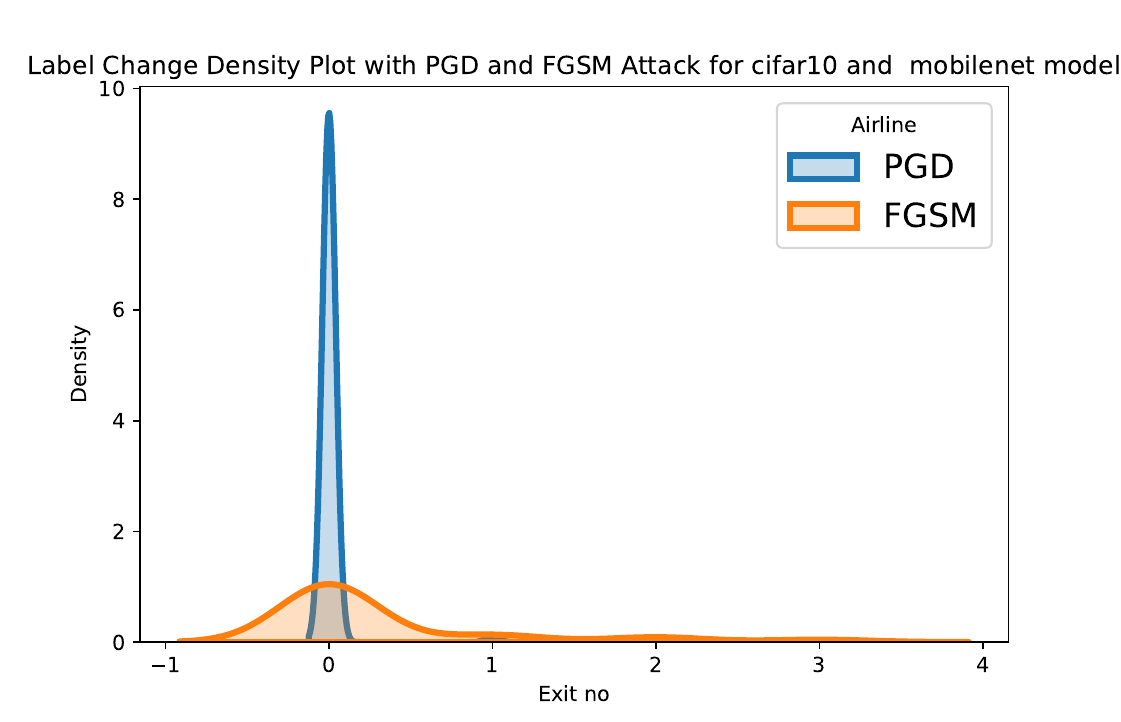}
        \caption{CIFAR-10 MobileNet}
    \end{subfigure}
    \begin{subfigure}[b]{0.285\textwidth}
        \includegraphics[width=\textwidth]{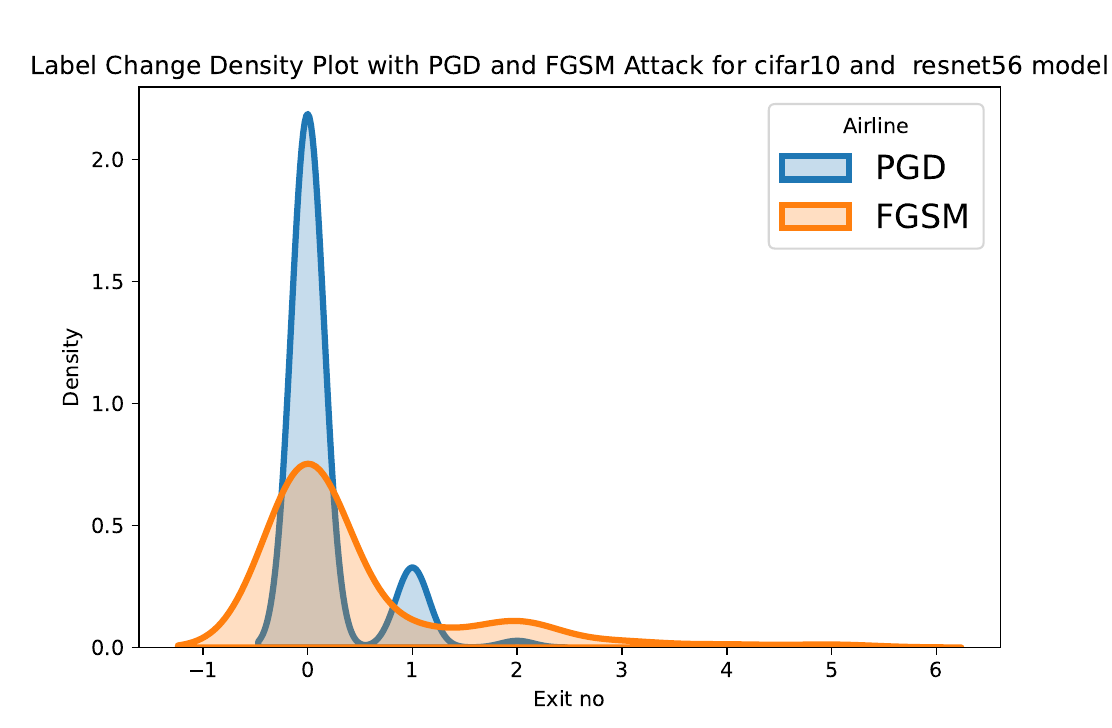}
        \caption{CIFAR-10 ResNet}
    \end{subfigure}
    \caption{Density plots representing during which exit number output label is changed because of PGD and FGSM attack (For CIFAR-10 data). The x axis represents the exit number while y axis represents the density.}
    \label{fig:rq3_1}
\end{figure*}

\begin{comment}
\begin{subfigure}[b]{0.28\textwidth}
        \includegraphics[width=\textwidth]{figs/c100_vgg_rq2_2.pdf}
        \caption{CIFAR-100 VGG}
    \end{subfigure}
    \begin{subfigure}[b]{0.292\textwidth}
        \includegraphics[width=\textwidth]{figs/c100_mnet_rq2_2.pdf}
        \caption{CIFAR-100 MobileNet}
    \end{subfigure}
    \begin{subfigure}[b]{0.29\textwidth}
        \includegraphics[width=\textwidth]{figs/c100_rnet_rq2_2.pdf}
        \caption{CIFAR-100 ResNet}
    \end{subfigure}

    \begin{subfigure}[b]{0.28\textwidth}
        \includegraphics[width=\textwidth]{figs/c100_vgg_rq2_2_bb.pdf}
        \caption{CIFAR-100 VGG}
    \end{subfigure}
    \begin{subfigure}[b]{0.29\textwidth}
        \includegraphics[width=\textwidth]{figs/c100_mnet_rq2_2_bb.pdf}
        \caption{CIFAR-100 MobileNet}
    \end{subfigure}
    \begin{subfigure}[b]{0.29\textwidth}
        \includegraphics[width=\textwidth]{figs/c100_rnet_rq2_2.pdf}
        \caption{CIFAR-100 ResNet}
    \end{subfigure}
\end{comment}

\begin{figure*}[t]
    \centering
    \begin{subfigure}[b]{0.28\textwidth}
        \includegraphics[width=\textwidth]{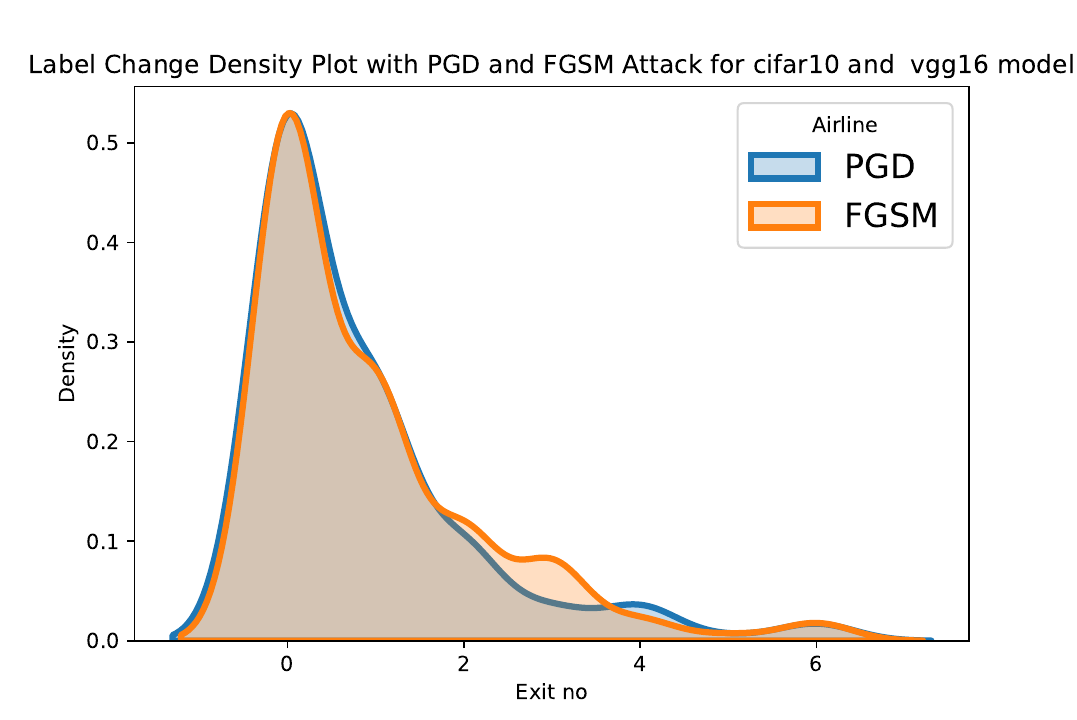}
        \caption{CIFAR-10 VGG}
    \end{subfigure}
    \begin{subfigure}[b]{0.29\textwidth}
        \includegraphics[width=\textwidth]{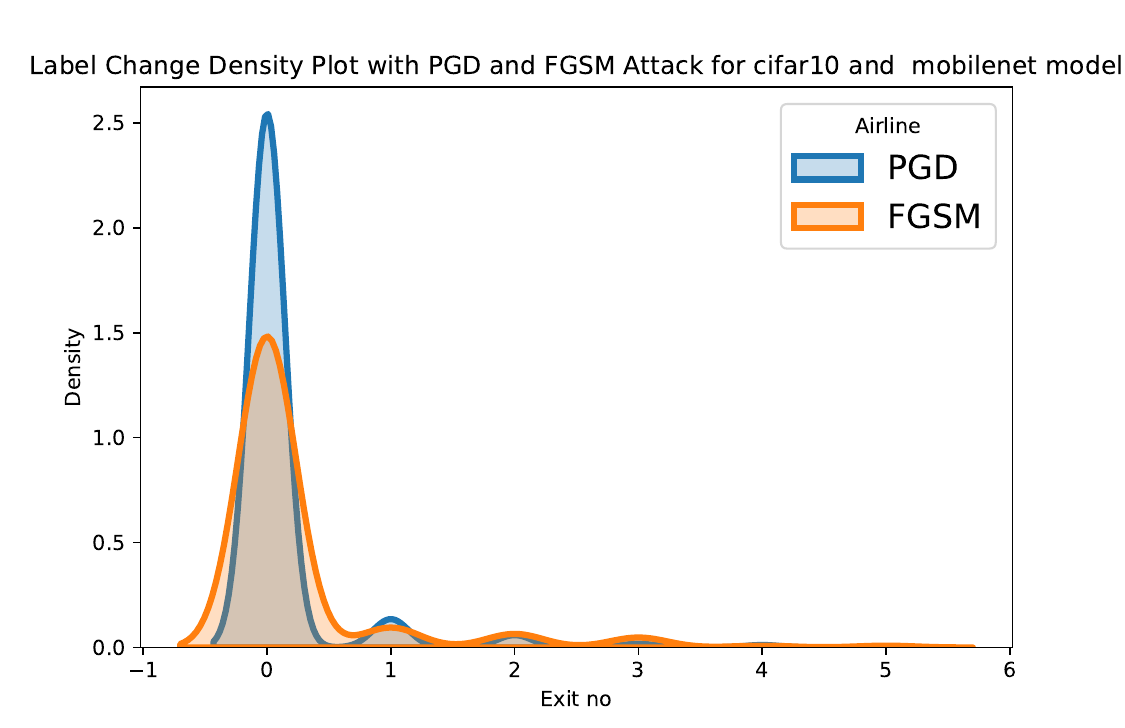}
        \caption{CIFAR-10 MobileNet}
    \end{subfigure}
    \begin{subfigure}[b]{0.28\textwidth}
        \includegraphics[width=\textwidth]{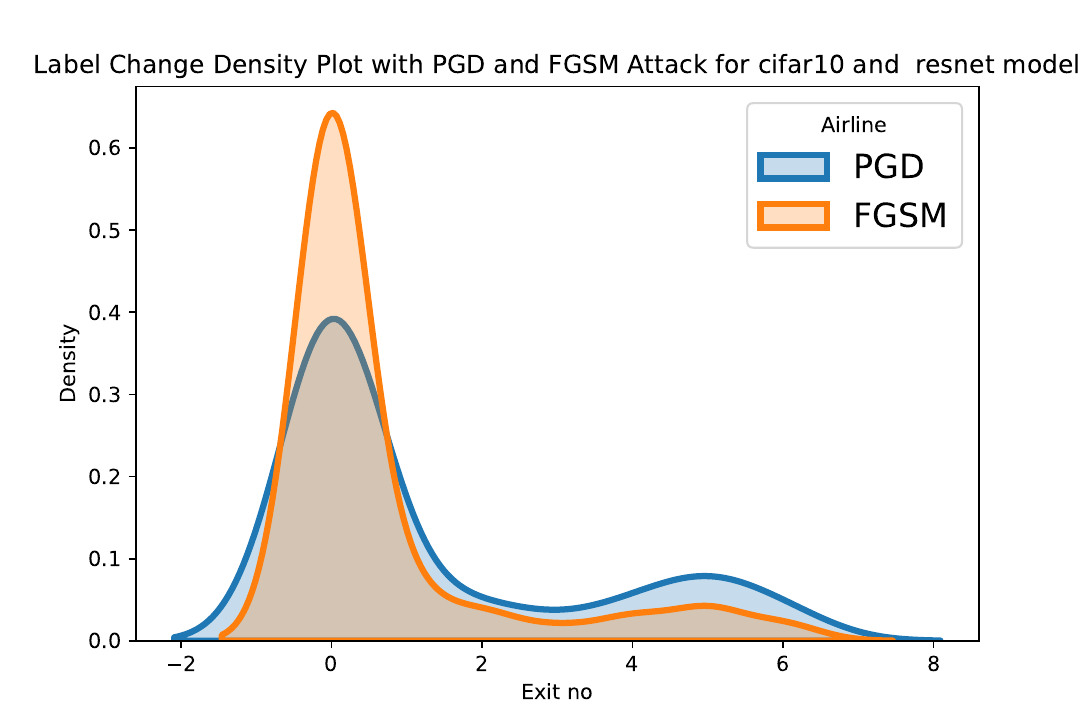}
        \caption{CIFAR-10 ResNet}
    \end{subfigure}
    
    \caption{Density plots representing during which exit number output label is changed because of PGD and FGSM \textbf{black-box} attack (For CIFAR-10 data). The x axis represents the exit number while y axis represents the density.}
    \label{fig:rq3_2}
\end{figure*}

%\section{Can we synthesize certain examples to evaluate the added attack surface of dynamic mechanism? }
\section{What design of DyNNs may impact the robustness against attack on dynamic mechanism?}
\label{sec:attack}

%reduce the effectiveness of the early-exits of DyNNs
%We design this study specifically to understand the extra attack surface introduced by the dynamic mechanisms.
In this section, we evaluate if specific examples can be generated only to understand the additional attack surface introduced by the dynamic mechanism in DyNNs and which design choices would be helpful in DyNNs to improve the robustness against this specific attack.
 First, we aim to design an attack such that the synthesized adversarial examples will not change the prediction of the final SDNN label, but change the prediction of all the early exits. 
This threat model is practical because the attack evades the existing detection that relies on the final output of SDNN while the attacker creates a situation where all the early exit networks do not provide correct prediction, therefore decreasing the usability of DyNNs. However, this attack is also challenging to be performed successfully because final layer output is dependent on the earlier exit layers and it is challenging to impact all the early exits' prediction without modifying the final exit's prediction.
%\begin{wrapfigure}{L}{0.35\textwidth}
  %\vspace{-2em}
  %\setlength{\textfloatsep}{-6pt}
    %\begin{minipage}{0.35\textwidth}
    %\centering
%\setlength\belo
\begin{algorithm}
\scriptsize
  \SetKwInOut{Input}{Inputs}\SetKwInOut{Output}{Outputs}
  %\SetKwInOut{Input}{Inputs}\SetKwInOut{Output}{Outputs}
  \caption{Input generation using Early Attack }\label{alg:L2}
  %\tcc{\textbf{forward crawling from a state}}
  \Input{$x:Input$ $Image$}
  \Output{$x':Perturbed$ $Image$
  }
  \BlankLine
  \Begin{
    Initialize($w$)\\
    $T=number\_of\_iterations$\\
    $iter\_no=0$\\
    \While{$iter\_no<T$}{
        $x'= \frac{\tanh(w) + 1}{2}$\\
        $output=model(x')$\\
        \If{${success}(output)$}{
        $return (x')$
        }
        $L = loss(x, w, c,\alpha)$\\
        $L_{new},w=Optim(L,w)$\\
        $iter\_no++$\\
    
    }
    $x'= \frac{\tanh(w) + 1}{2}$\\
}
\end{algorithm} 
\setlength{\floatsep}{-6pt}

%\end{minipage}
%\setlength\belowdisplayskip{0pt}
%\vspace{-2em}
  %\end{wrapfigure}

\subsection{Problem Formulation} We propose a novel attack technique called \textit{Early Attack} to evaluate the effectiveness of the early layers of DyNNs. Let's assume $f$ is an DyNN with $N$ exits. Given an input $x$, the output softmax layer in an exit $i$ can be defined as $y^{i}=f_i(x)$, where $i=1,2,3...N$. For synthesizing adversarial examples, we have two main objectives. First, the initial prediction in the $N^{th}$ exit (final layer) does not change. Let's assume, the initial prediction at final layer is $p$. Second, the prediction of all the other exits should be different than $p$.

Based on the aforementioned objectives, we can propose an iterative optimization procedure to optimise a loss function $L$. For each of the objectives, one loss function component is formulized.
For the first objective, we propose loss function $L1=(-1*\sum_{j \neq p}max(y^{N}_p-y^{N}_j,0))$. In $L1$, we maximize the difference between softmax value of label $p$ and other label's softmax value, therefore the prediction won't get changed at the final exit.
For the second objective, we propose loss function $L2=(\sum_{i = 1}^{N}y^{i}_p)$. The $L2$ ensures that for any other exit than the final exit, $p$'s softmax value would be minimized. The final loss function $L=\alpha * L1+ L2$. Here the $\alpha$ is a user-defined variable that provides balance between two loss terms.

Finally, we need to ensure the added perturbation to generate adversarial examples are limited, hence, the final optimization function would be, $minimize(||\delta||+c \cdot L)$, where, $(x+\delta) \in [0,1]^n$.
Here, $\delta$ represents the added perturbation and $c$ is a user-defined variable to provide weightage on a specific component.
$c$ controls the magnitude of generated perturbation ($||\delta_i||$), where a large $c$ makes the loss function more dependant on the $L$. 

%Whereas a smaller $c$ makes the loss function more dependant on $||\delta_i||$, allowing only smaller perturbation.

This constrained optimization problem in $\delta$ can be converted into a non-constrained optimization problem in $w$, where the relationship between $\delta$ and $w$ is: $ \delta = \frac{\tanh(w) + 1}{2} - x$
The $tanh$ function would ensure that the generated adversarial input values stay between 0 and 1. The equivalent optimization problem in $w$ is:
\begin{gather}\label{eq:3}
\underset{w}{\textrm{minimize}} \;  \Bigg|\Bigg|\frac{\tanh(w) + 1}{2} - x\Bigg|\Bigg| + c \cdot L
\end{gather}

Algorithm \ref{alg:L2} shows the optimization algorithm. The algorithm outputs the adversarial input $x'$ given a benign image $x$ as input. $w$ is initialized to a random tensor that has equal shape as the input image.
For each iteration, the loss function of the attack is computed (at line 11). Based on the back-propagated loss, the optimizer updates $w$ with its next value. Once the iteration threshold ($T$) is reached or the attack is successful, the algorithm computes and returns the adversarial input $x'$ (at Line 9 and 15). 

%Detailed iterative optimization can be found in Appendix (\ref{sec:algo}).

%\lipsum[1]
  
  %\lipsum

\subsection{Evaluation}

\subsubsection{Evaluation SetUp.}

%\textbf{Datasets and Models.} We use the same dataset and model setup as previous RQs.

\textbf{Baseline.} We use PGD and FGSM attacks as baseline to modify the early exit label. 

\textbf{Metric.} We use attack success rate as metric in the evaluation. If the adversarial input generated final layer output is same as the output generated by benign input and all the other exit layers output is different than the final output label, then we consider attack is successful for that particular adversarial input.  

\textbf{Hyperparameters.} We use $\alpha=\{0.001,0.01,0.1,1,20,40\}$ and $c=50$ as hyperparameters.

%use six values of $\alpha$ to evaluate this section: 0.001,0.01,0.1,1,20,40. We consider $c$ value as 50.

\subsubsection{Effectiveness}

Table \ref{tab:HP} shows the evaluation results of the attack success rate of Early Attack and baseline techniques.  Except two model-dataset pair, Early Attack's success rate is higher than 80\% in all other scenarios. PGD and FGSM attacks are unsuccessful because the loss function only considers early layer output and does not consider final layer output. Although FGSM's attack success rate is higher than PGD's success rate. Also, w.r.t best performing $\alpha$ values, $\alpha=1$ performs well for 70\% of the model-dataset pairs.

We also analyze why early attack fails for VGG16 and MobileNet models for CIFAR-10 data. First, for CIFAR-10 data, final fully connected layer has lower number of parameters than the final layer for CIFAR-100 data. Hence, the dependency between the final layer and earlier layers is higher for CIFAR-10. However, for CIFAR-10 data, attack success rate is higher for ResNet model. First we discuss the failed cases for VGG16 and MobileNet model, and then we discuss why the attack succeeded for ResNet model.

For VGG16 model, the failed examples can be divided into two types. For the first type, the final output label is changed with all the other early exit layers. In the second type, first few layers get the output label mis-classified, but along with the final layer, few previous layers also get the output correctly classified. For MobileNet, all the layers except the final and last to final layer get the output label mis-classified. For MobileNet, we also find that the final two exits are separated by only two layers, which is significantly lower than other model's separation layers between final two exits. Hence, the dependency between final two layers is higher.

For ResNet model, there are 56 layers divided into 27 blocks. The exits are sparsely divided between these blocks. For Mobilenet and VGG models, the exit distributions are less sparse. Hence, the dependency between exits is lower for ResNet and because of that reason, we could successfully attack ResNet model.

\begin{table}[h]
%\begin{wraptable}{r}{6.5cm}
\centering
%\adjustbox{max width=0.35\textwidth,center}{
\scriptsize
\setlength{\tabcolsep}{0.3em}
{\renewcommand{\arraystretch}{1}
\caption{ \textbf{Attack accuracy percentage of Early Attack and the baseline techniques against different model and dataset, along with $\alpha$ value.} }
    \label{tab:HP}
\begin{tabular}{|*{56}{c|}}
\hline
Dataset &Model & Early Attack & best $\alpha$ val &  PGD & FGSM \\
\hline
\multirow{3}{*}{CIFAR-10} 
  & VGG & \textbf{35} &1 & 0 & 0 \\
\cline{2-6} & MobileNet & \textbf{11} &0.1 & 0 & 0 \\
\cline{2-6} &ResNet & \textbf{81} &1&   0 & 0\\
\hline
\multirow{3}{*}{CIFAR-100} 
 & VGG & \textbf{86} &20&   0 & 1 \\
\cline{2-6} & MobileNet & \textbf{97} &1&   0 & 0 \\
\cline{2-6} &ResNet & \textbf{96} &1&   0 & 2\\
\hline
\end{tabular}}
%\end{wraptable}
\end{table}

\subsubsection{Transferability} In this section, we discuss about the transferability of the Early Attack examples. As Early Attack has two different components, instead of measuring attack success rate directly, we measure two parameters $T1$ and $T2$. $T1$ represents the percentage of inputs for which the final output remains same as the output generated by benign input. $T2$ represents from the examples selected from $T1$, how many early exit layers on average are mis-classified. Having both high $T1$ and $T2$ values ensures transferability. 

\begin{table}[hbtp]
%\vspace{-4mm}
\centering
%\scalebox{0.8}{
%\adjustbox{width=0.35\textwidth,center}{
\scriptsize
\setlength{\tabcolsep}{0.3em}
\caption{\textbf{$T1$ and $T2$ values for measuring transferability between three models. \textit{TM} represents Target Model and \textit{SM} represents Surrogate Model. $T1$ presents the percentage of inputs for which the final output remains same as the output generated by benign input. $T2$ represents from the examples selected from $T1$, how many early exit layers on average is mis-classified.}}
    \label{tab:ITP}
{\renewcommand{\arraystretch}{1}
 \begin{tabular}{|l| c| c|c| c|c|c|c|}

 \cline{1-8}& &\multicolumn{3}{c|}{CIFAR-10}& \multicolumn{3}{c|}{CIFAR-100}\\
 
 \hline
 
 Type & \diagbox{SM}{\rule{5mm}{0mm}TM} & VGG & MNet &RNet & VGG & MNet &RNet  \\  
 \hline
 \multirow{3}{*}{$T1$} & VGG & -- &  85\% & 73\%& -- & 65\% & 39\%\\ 
 \cline{2-8}& MNet & 86\% &  -- &74\% &51\% & --& 38\%\\
 \cline{2-8}& RNet & 89\% &  82\% &-- &79\% &75\% &--\\
 
 \hline
 
 \multirow{3}{*}{$T2$} & VGG & -- & 0.73  & 1.65& -- & 1.58 & 2.69 \\ 
 \cline{2-8}& MNet & 1.06 &  -- &2.3 &1.52 &-- & 3.26\\
 \cline{2-8}& RNet & 0.95 &  0.85 &-- &1.32 &1.28 &--\\
 
 \hline
 
\end{tabular}

}

%}

\end{table}    

We show the transferability results in Table \ref{tab:ITP}. For CIFAR-10 data, the $T1$ values are high, but $T2$ values are significantly low except for MobileNet to ResNet transferability. For CIFAR-100 data, $T2$ values are higher than of CIFAR-10 data, however, $T1$ values are low. From the results, we can notice that the generated examples either can keep the final layer label the same or can change the early exit layer outputs. 
Our finding suggests that early attack transferability is limited.

%First we measure

\begin{center}
\begin{tcolorbox}[colback=gray!10,%gray background
                  colframe=black,% black frame colour
                  width=8cm,% Use 8cm total width,
                  arc=1mm, auto outer arc,
                  boxrule=0.9pt,
                 ]
 \textbf{Finding 9}:  \textit{With increasing number of exits in DyNNs, the dependency between multiple exits will increase. Hence, more exits in DyNNs can increase the robustness against the \textit{Early Attack}.}
 
 \textbf{Finding 10}:  \textit{\textit{Early Attack} transferability  between DyNNs is not significant.}
 
 %\textbf{Finding 2}:  \textit{Having lower number of parameters between }

\end{tcolorbox}
\end{center}

\section{Conclusion} In this work \footnote{https://github.com/anonymous2015258/Early\_Attack/tree/main}, we discuss the robustness of including dynamic mechanism in DNN through three research questions. We find out that DyNNs are more robust than SDNNs and also efficient to generate adversarial examples. We also propose DyNN design choices through final two RQs. Finally, we propose a novel attack to understand additional attack space in DyNNs.

\newpage
%\bibliography{iclr2023_conference}
%\bibliographystyle{iclr2023_conference}
%\bibliographystyle{icml2023}
{\small
\bibliographystyle{ieee_fullname}
\bibliography{iclr2023_conference}
}

\clearpage
\appendix
\appendixpage
%\section{Appendix}

\section{RQ1 Density Plots} 
\begin{figure*}[t]
    \centering
    \begin{subfigure}[b]{0.28\textwidth}
        \includegraphics[width=\textwidth]{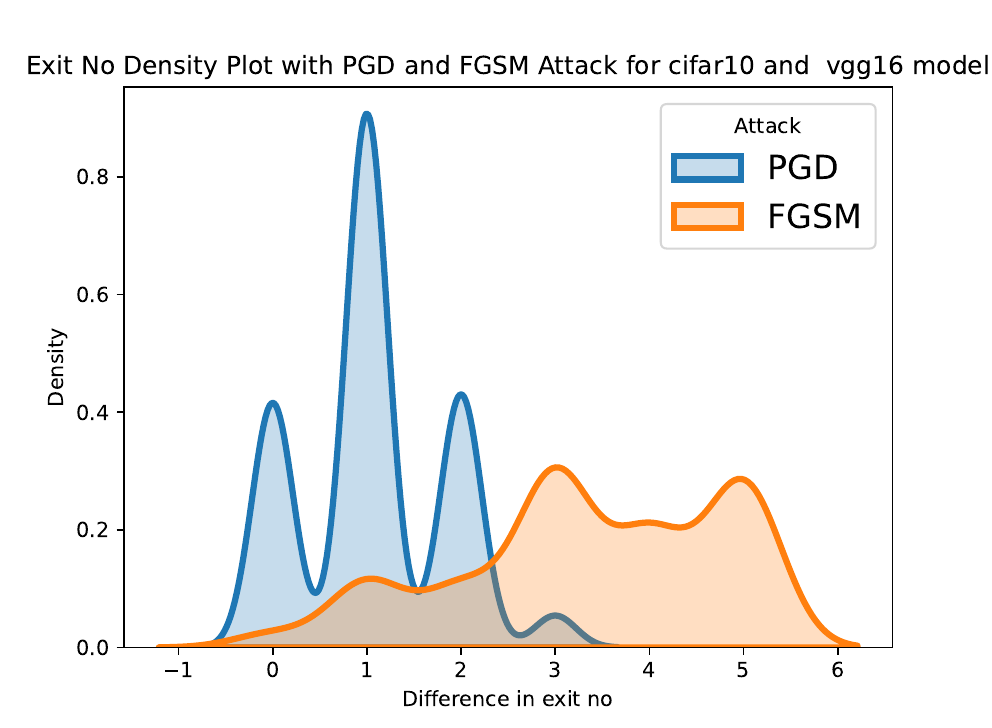}
        \caption{CIFAR-10 VGG}
    \end{subfigure}
    \begin{subfigure}[b]{0.29\textwidth}
        \includegraphics[width=\textwidth]{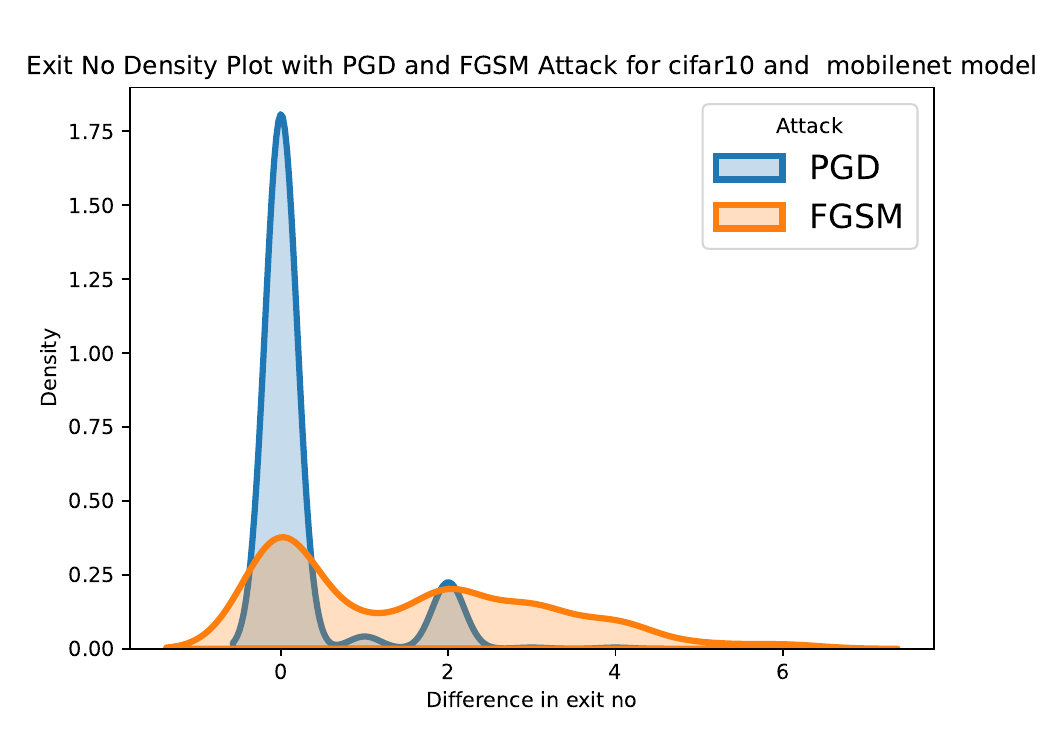}
        \caption{CIFAR-10 MobileNet}
    \end{subfigure}
    \begin{subfigure}[b]{0.28\textwidth}
        \includegraphics[width=\textwidth]{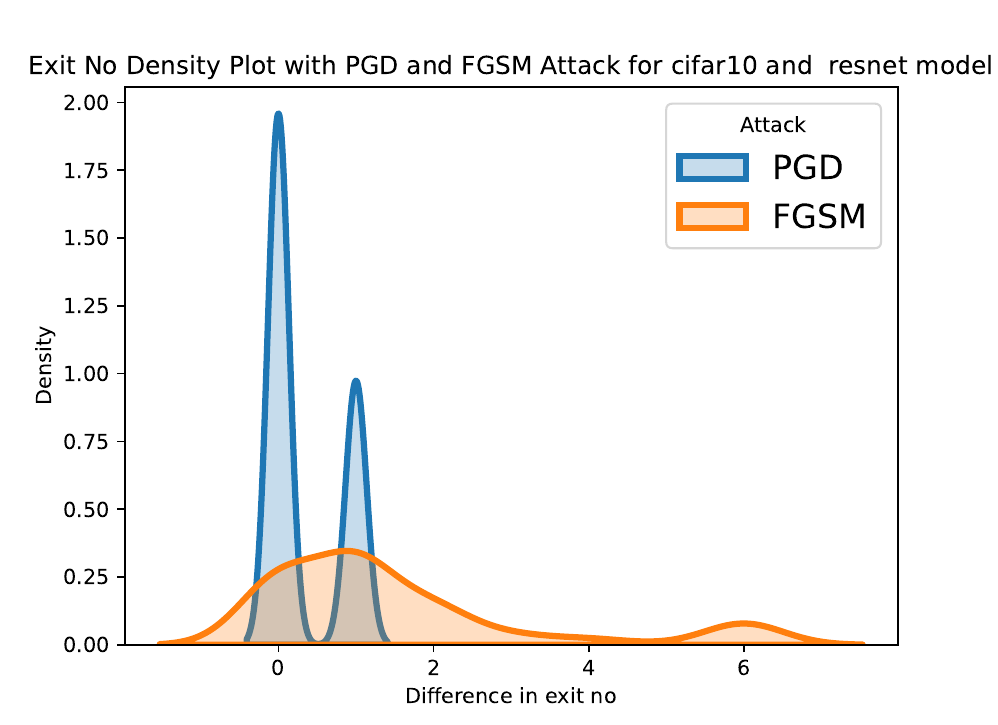}
        \caption{CIFAR-10 ResNet}
    \end{subfigure}
    \begin{subfigure}[b]{0.28\textwidth}
        \includegraphics[width=\textwidth]{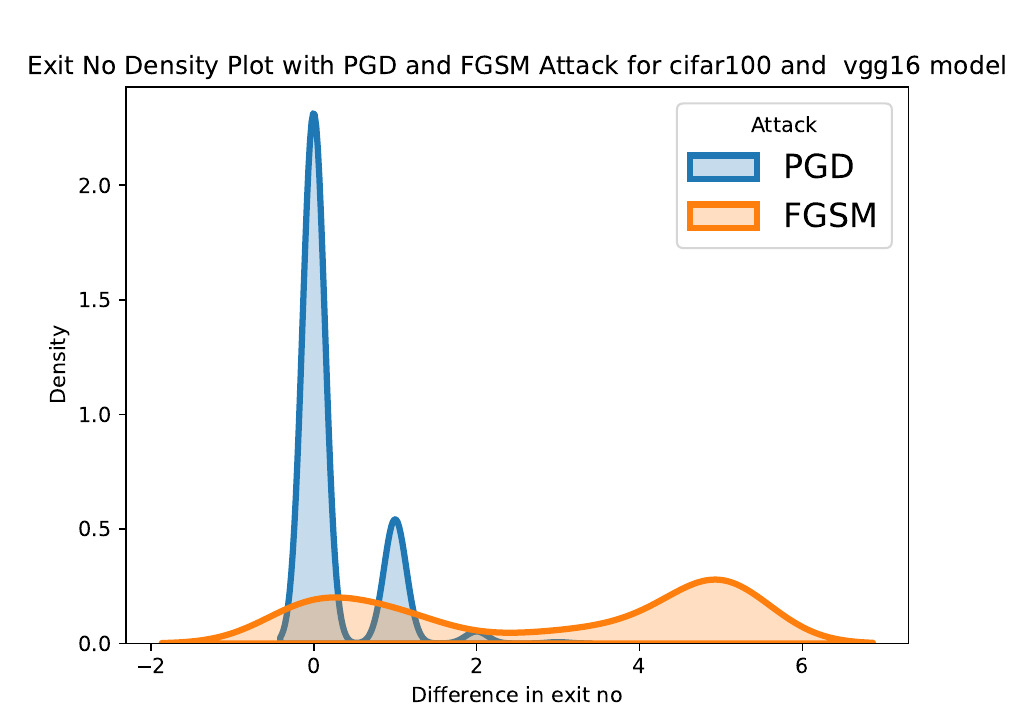}
        \caption{CIFAR-100 VGG}
    \end{subfigure}
    \begin{subfigure}[b]{0.29\textwidth}
        \includegraphics[width=\textwidth]{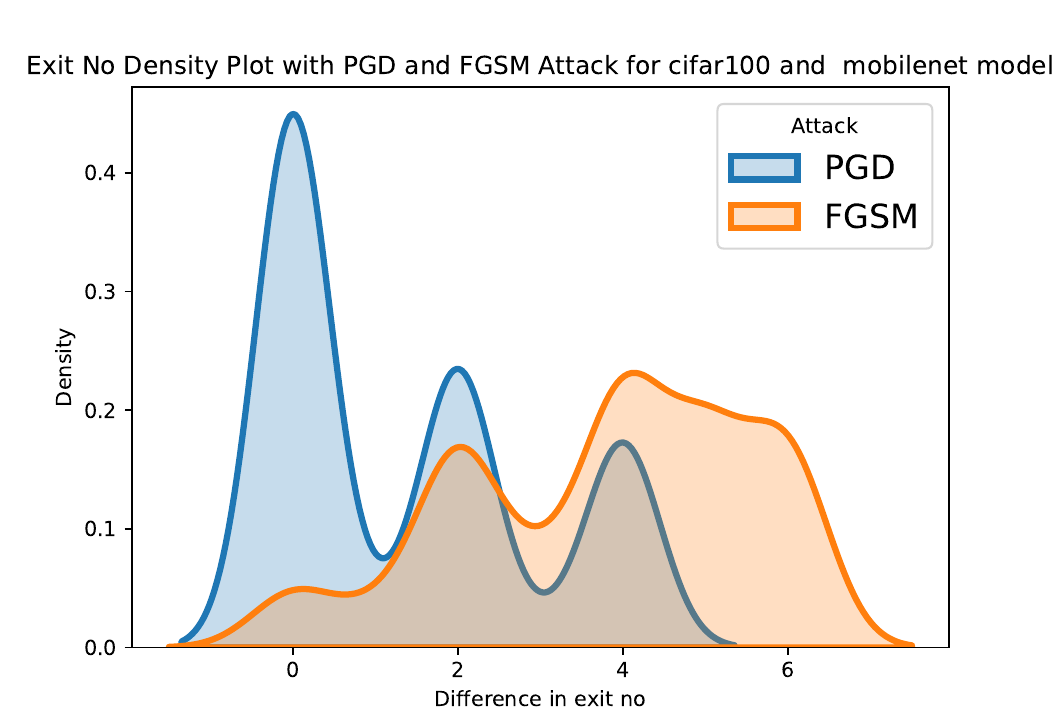}
        \caption{CIFAR-100 MobileNet}
    \end{subfigure}
    \begin{subfigure}[b]{0.28\textwidth}
        \includegraphics[width=\textwidth]{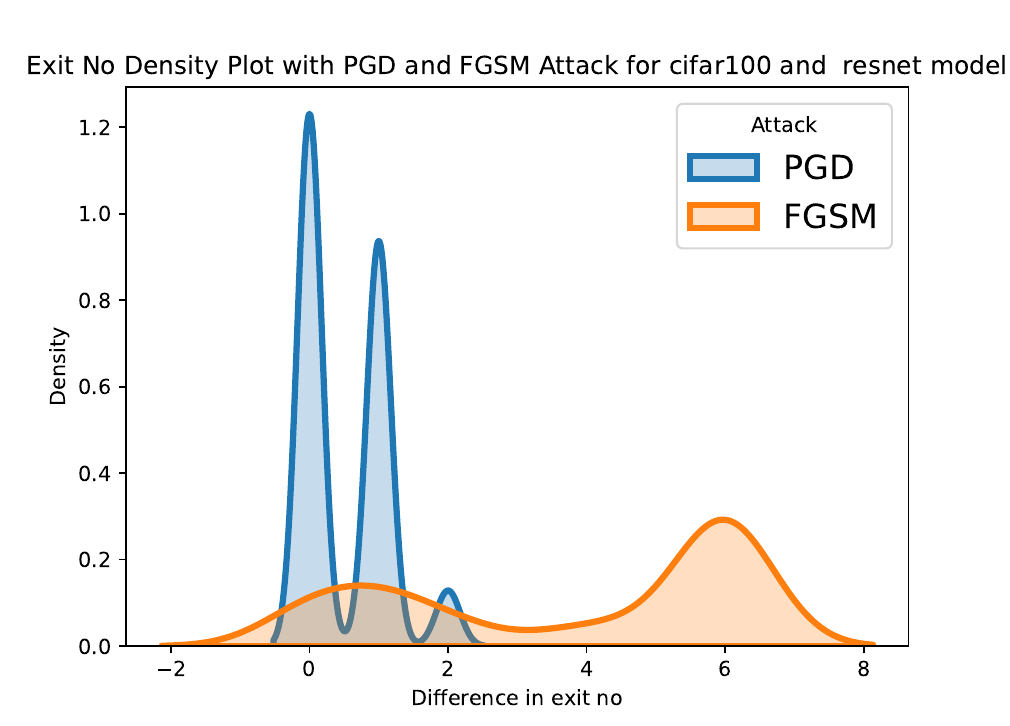}
        \caption{CIFAR-100 ResNet}
    \end{subfigure}
    \caption{Density plots of exit numbers in DyNN that are used to generate black-box adversarial inputs using PGD and FGSM attacks. The x axis represents the exit number while y axis represents the density.}
    \label{fig:rq1_2}
\end{figure*}

The density plots in Figure \ref{fig:rq1_2} shows the probability density plots of exit numbers in DyNN that are used to generate black-box adversarial inputs using PGD and FGSM attacks.

\begin{figure*}[t]
    \centering
    \begin{subfigure}[b]{0.28\textwidth}
        \includegraphics[width=\textwidth]{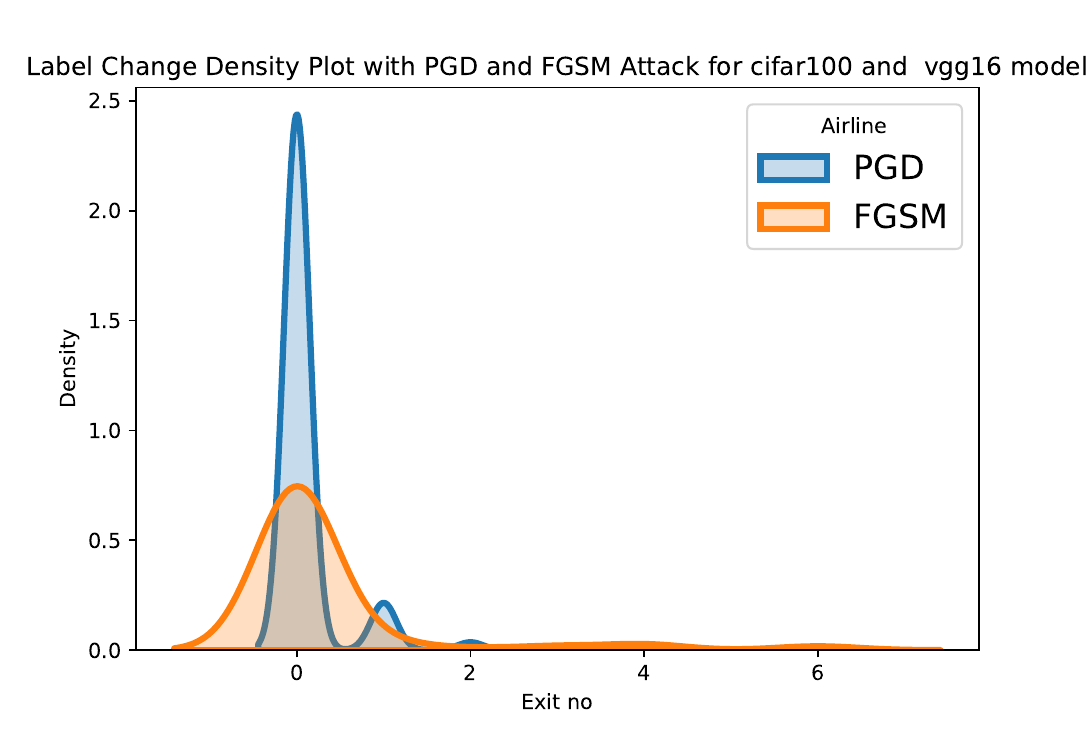}
        \caption{CIFAR-100 VGG}
    \end{subfigure}
    \begin{subfigure}[b]{0.292\textwidth}
        \includegraphics[width=\textwidth]{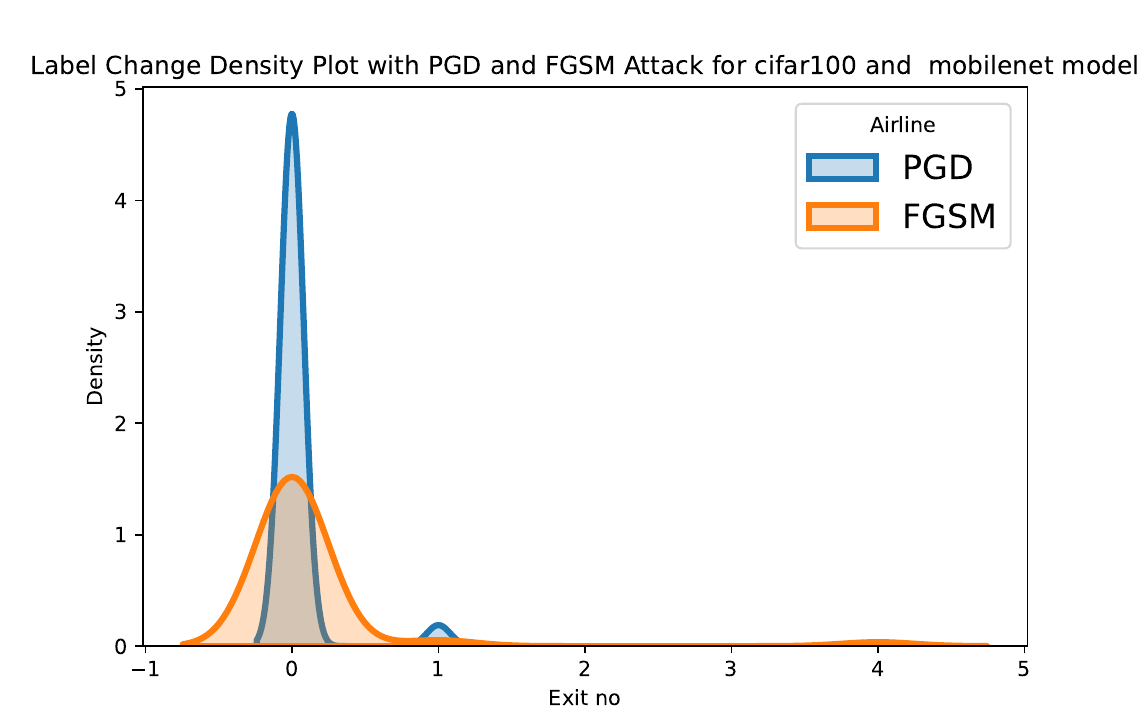}
        \caption{CIFAR-100 MobileNet}
    \end{subfigure}
    \begin{subfigure}[b]{0.29\textwidth}
        \includegraphics[width=\textwidth]{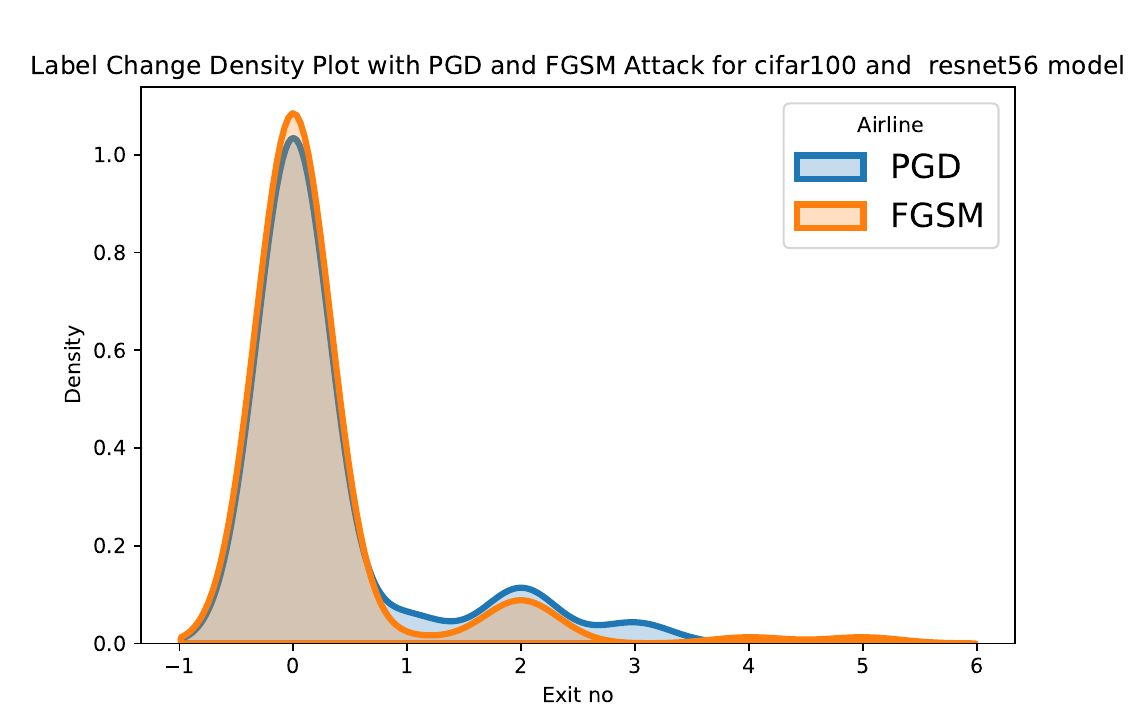}
        \caption{CIFAR-100 ResNet}
    \end{subfigure}
    \caption{Density plots representing during which exit number output label is changed because of PGD and FGSM attack (For CIFAR-100 data). The x axis represents the exit number while y axis represents the density.}
    \label{fig:rq3_1_appnd}
\end{figure*}

\begin{figure*}[t]
    \centering
    \begin{subfigure}[b]{0.28\textwidth}
        \includegraphics[width=\textwidth]{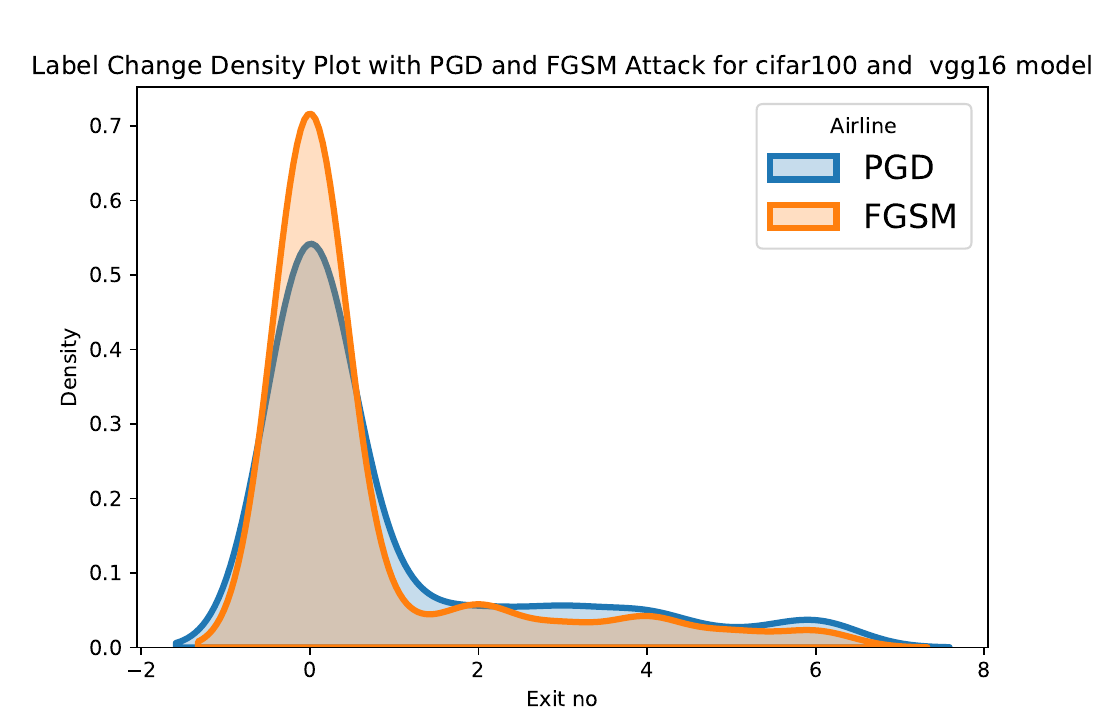}
        \caption{CIFAR-100 VGG}
    \end{subfigure}
    \begin{subfigure}[b]{0.29\textwidth}
        \includegraphics[width=\textwidth]{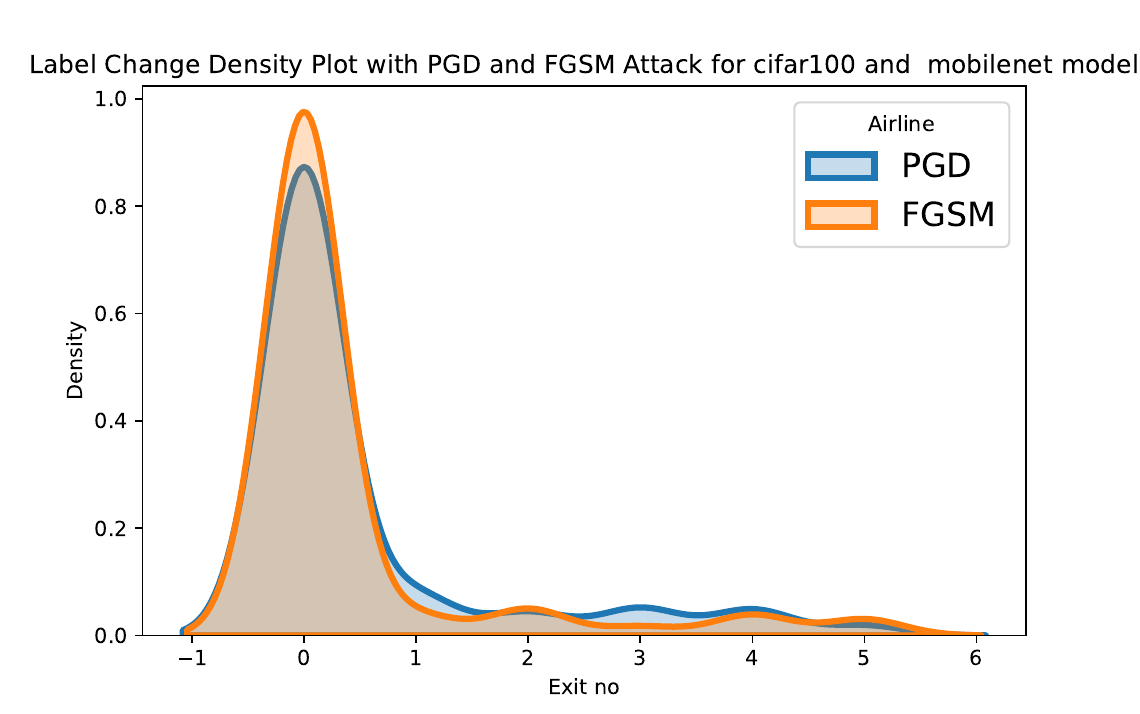}
        \caption{CIFAR-100 MobileNet}
    \end{subfigure}
    \begin{subfigure}[b]{0.29\textwidth}
        \includegraphics[width=\textwidth]{figs/c100_rnet_rq2_2.pdf}
        \caption{CIFAR-100 ResNet}
    \end{subfigure}
    \caption{Density plots representing during which exit number output label is changed because of PGD and FGSM \textbf{black-box} attack (For CIFAR-100 data). The x axis represents the exit number while y axis represents the density.}
    \label{fig:rq3_2_append}
\end{figure*}

\begin{figure*}[t]
    \centering
    \begin{subfigure}[b]{0.28\textwidth}
        \includegraphics[width=\textwidth]{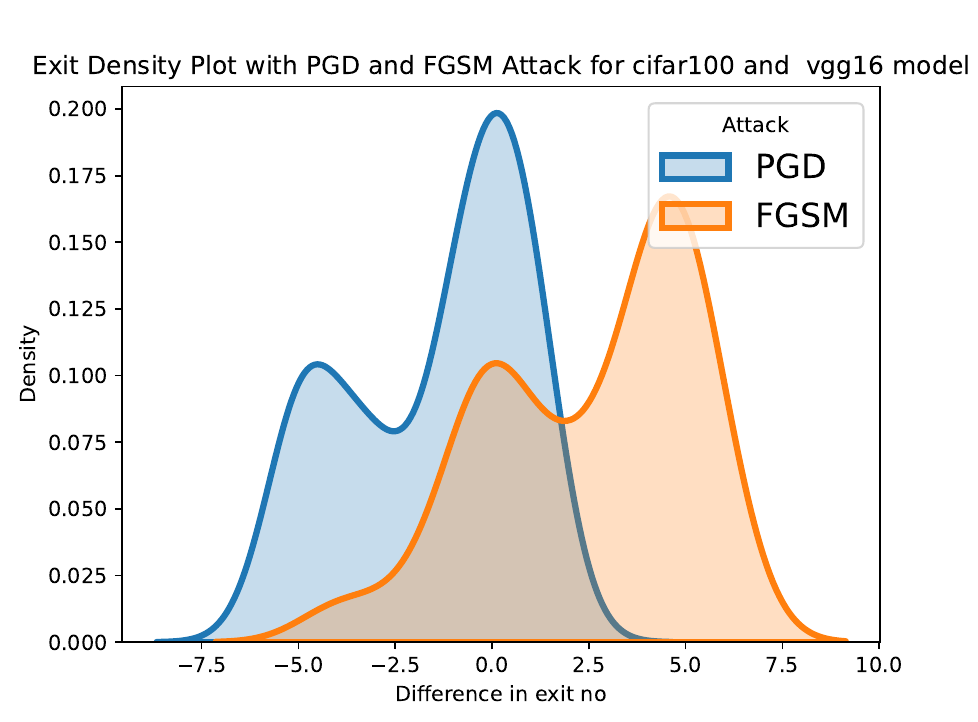}
        \caption{CIFAR-100 VGG}
    \end{subfigure}
    \begin{subfigure}[b]{0.29\textwidth}
        \includegraphics[width=\textwidth]{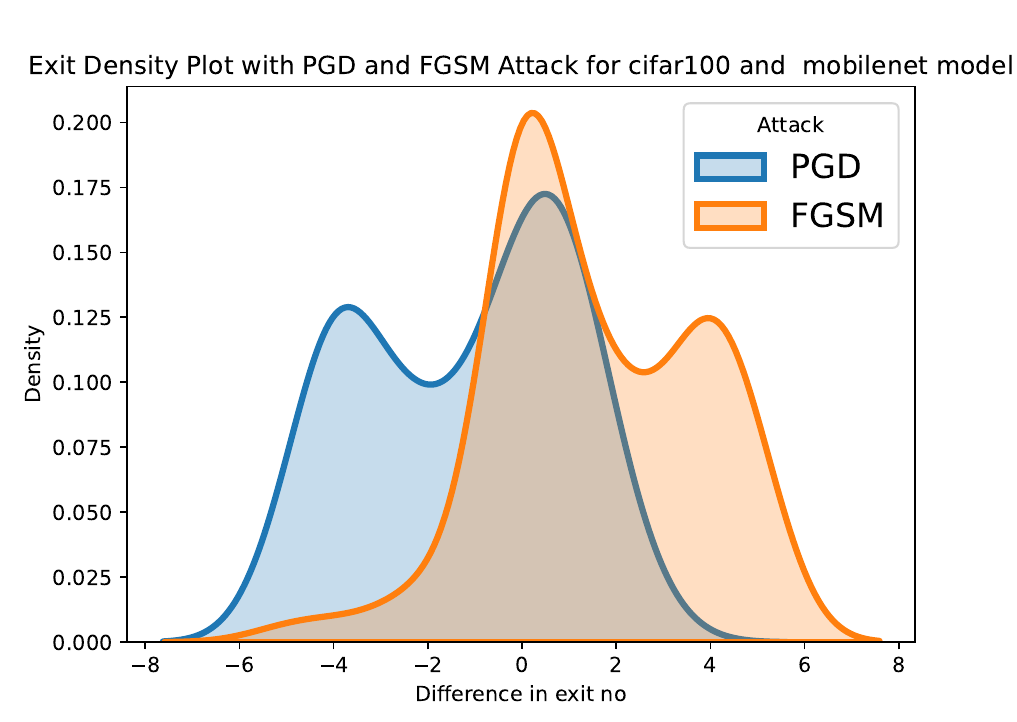}
        \caption{CIFAR-100 MobileNet}
    \end{subfigure}
    \begin{subfigure}[b]{0.29\textwidth}
        \includegraphics[width=\textwidth]{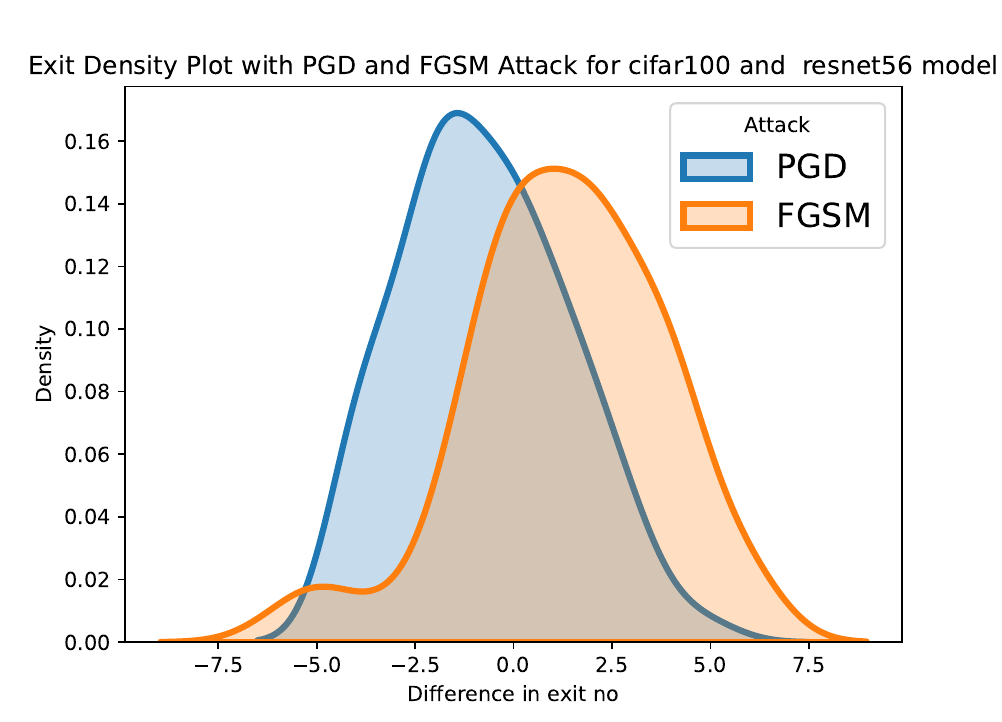}
        \caption{CIFAR-100 ResNet}
    \end{subfigure}
    \caption{Density plots of change in exit numbers because of PGD and FGSM attack (For CIFAR-100 data). The x axis represents the change in exit number while y axis represents the density.}
    \label{fig:rq2_1_appnd}
\end{figure*}

\begin{figure*}[t]
    \centering
    \begin{subfigure}[b]{0.28\textwidth}
        \includegraphics[width=\textwidth]{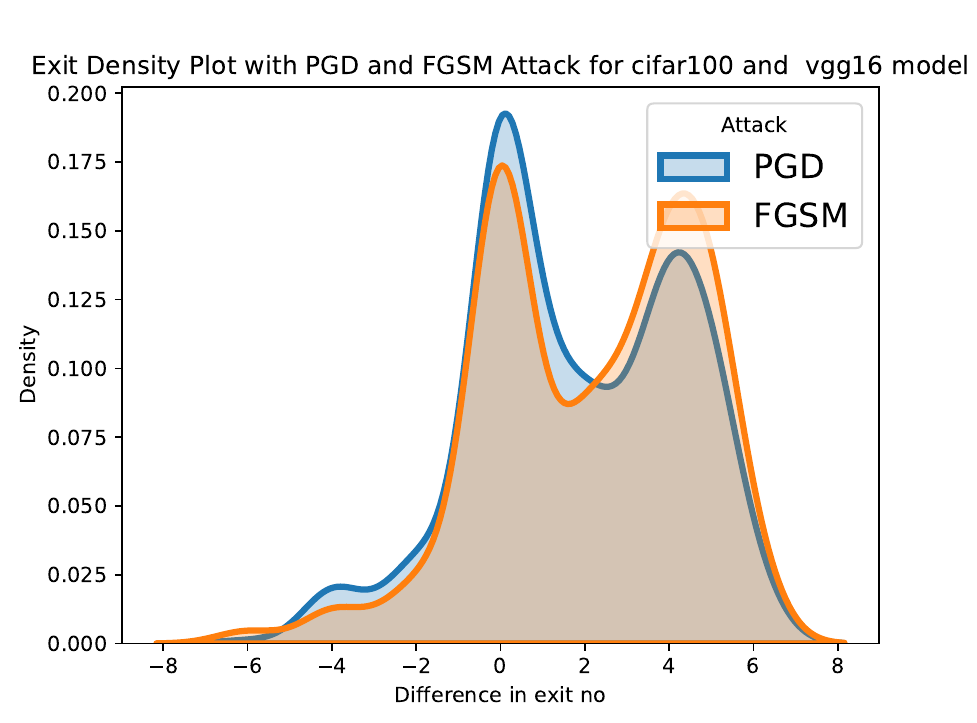}
        \caption{CIFAR-100 VGG}
    \end{subfigure}
    \begin{subfigure}[b]{0.29\textwidth}
        \includegraphics[width=\textwidth]{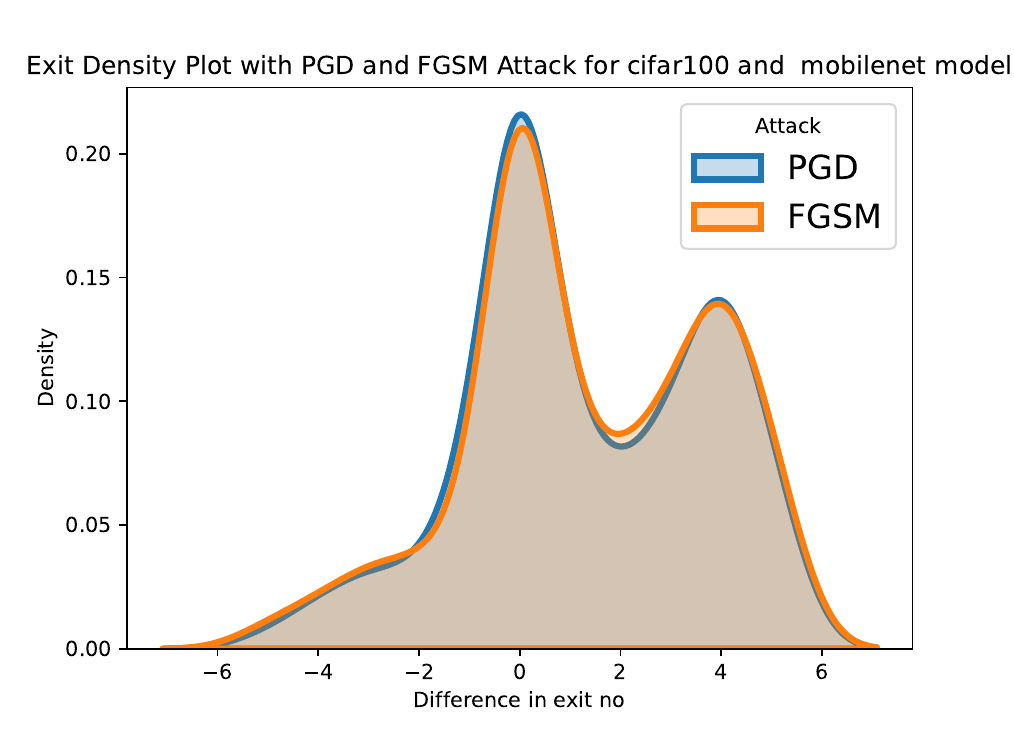}
        \caption{CIFAR-100 MobileNet}
    \end{subfigure}
    \begin{subfigure}[b]{0.28\textwidth}
        \includegraphics[width=\textwidth]{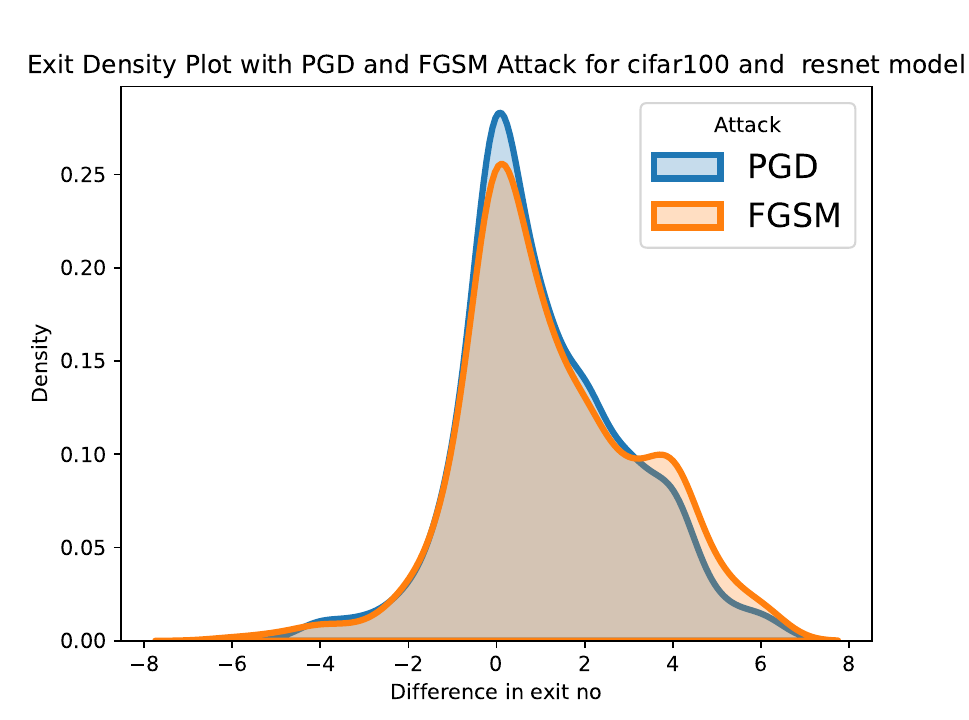}
        \caption{CIFAR-100 ResNet}
    \end{subfigure}
    \caption{Density plots of change in exit numbers because of PGD and FGSM \textbf{black-box} attack (For CIFAR-100 data). The x axis represents the change in exit number while y axis represents the density.}
    \label{fig:rq2_2_appnd}
\end{figure*}

\section{Early Attack Success Rate with Different $\alpha$ values}

\begin{table*}[h]
\centering
%\adjustbox{max width=0.35\textwidth,center}{
%\scriptsize
%\setlength{\tabcolsep}{0.3em}
{\renewcommand{\arraystretch}{1}
\caption{ \textbf{Attack accuracy percentage of Early Attack with different $\alpha$ values} }
    \label{tab:HP2}
\begin{tabular}{|*{56}{c|}}
\hline
Dataset &Model & EA($\alpha$=0.001) & EA($\alpha$=0.01) &  EA($\alpha$=0.1) & EA($\alpha$=1) & EA($\alpha$=20) & EA($\alpha$=40) \\
\hline
\multirow{3}{*}{CIFAR-10} 
  & VGG & 0 &0 & 0 & 35 & 10&4 \\
\cline{2-8} & MobileNet & 0 &0 & 11 & 4 & 0&0 \\
\cline{2-8} &ResNet & 7 &32&   73 & 81 &49 &32\\
\hline
\multirow{3}{*}{CIFAR-100} 
 & VGG & 1 &5&   48 & 82 &86 &70 \\
\cline{2-8} & MobileNet & 0 &16&   74 & 97 &92 &77\\
\cline{2-8} &ResNet & 46 &74&   92 & 96 &96 &93\\
\hline
\end{tabular}}
\end{table*}

\section{Transferability}
\label{sec:trans}

Table \ref{tab:HP2} shows the $T1$ and $T2$ values of all three models on two datasets.

\section{RQ1 results based on MI-FGSM attack}

Here, through Figure \ref{fig:mifgsm}, we show the adversarial transferability results between DyNNs and SDNNs using MI-FGSM attack. These results again confirm that adversarial examples from DyNN to SDNN are more transferable than adversarial examples from SDNN to DyNN.

\section{Transferability experiments on MI-FGSM attack}

\begin{figure*}[t]
    \centering
    \begin{subfigure}[b]{0.465\textwidth}
        \includegraphics[width=\textwidth]{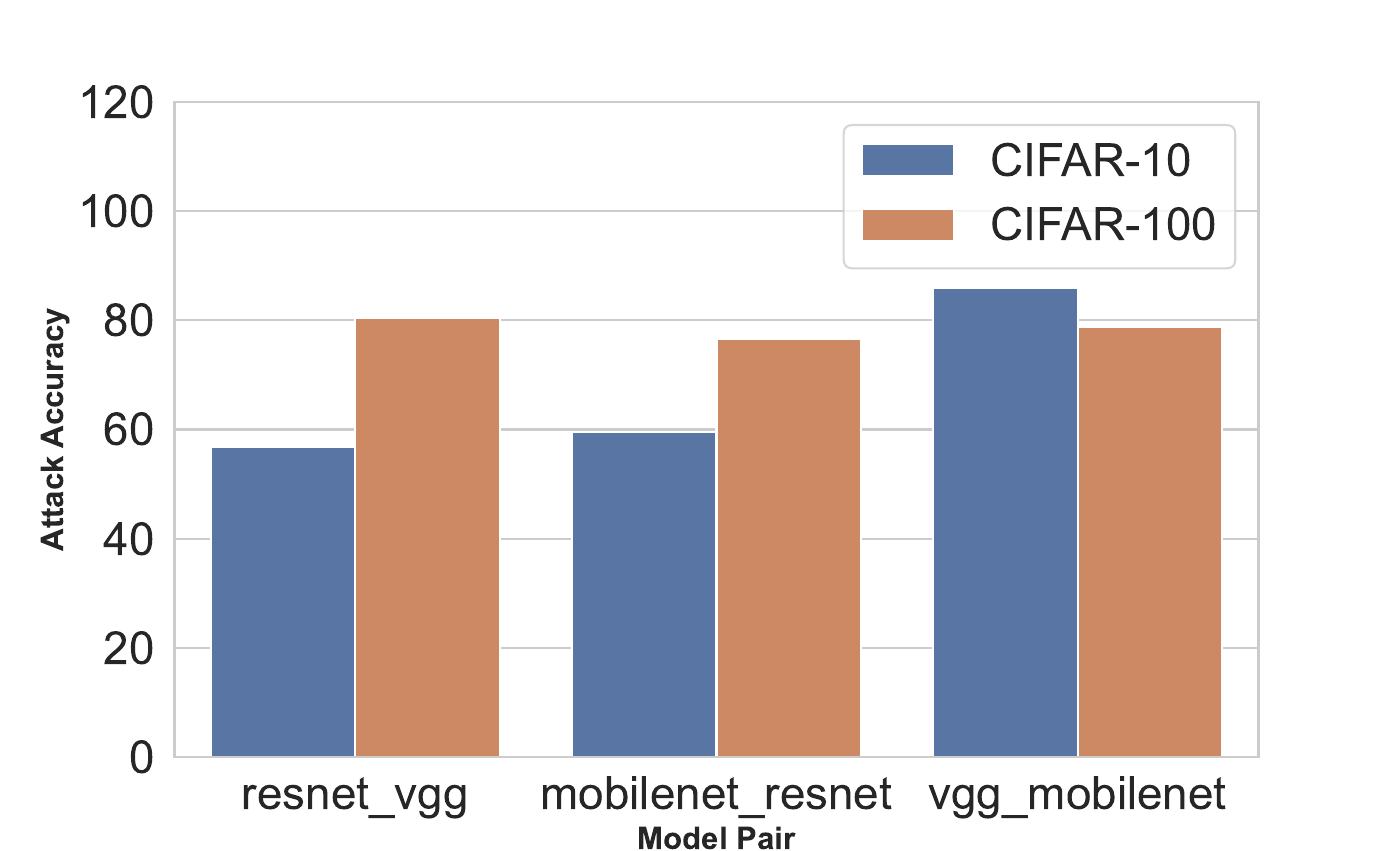}
        \caption{Target DyNN MI-FGSM}
    \end{subfigure}
    \begin{subfigure}[b]{0.45\textwidth}
        \includegraphics[width=\textwidth]{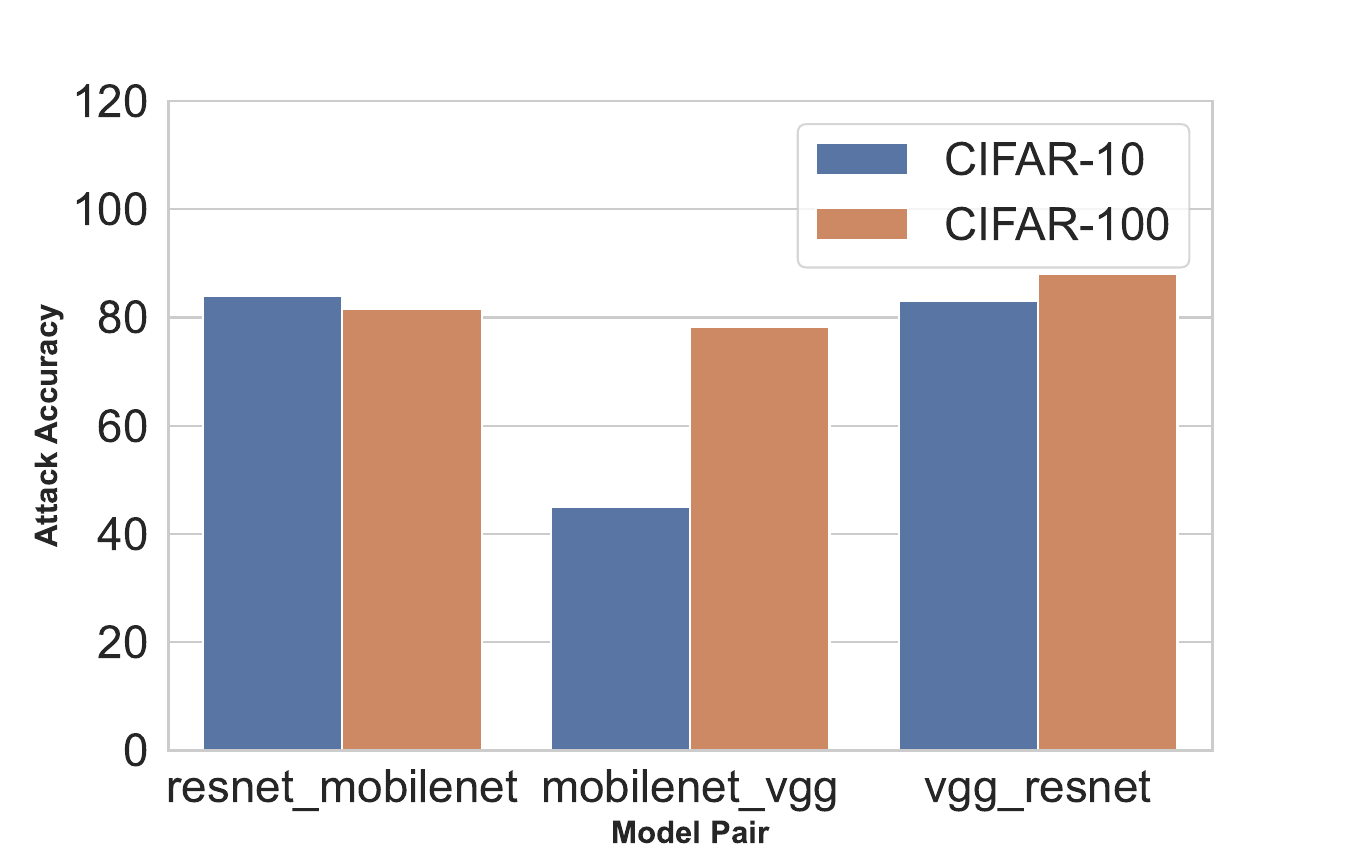}
        \caption{Target SDNN MI-FGSM}
    \end{subfigure}
    \caption{Transferable Attack Success Rate for MI-FGSM attack}
    \label{fig:mifgsm}
\end{figure*}

Through Figure \ref{fig:mifgsm}, we show the S2D and D2S transferability with MI-FGSM attack~\cite{dong2018boosting}. The results confirm our claim that D2S transferability is higher than S2D transferability.

\section{Comparing different adversarial images}

\begin{figure*}[t]
    \centering
    \begin{subfigure}[b]{0.15\textwidth}
        \includegraphics[width=\textwidth]{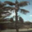}
        %\caption{CIFAR-10 VGG}
    \end{subfigure}
    \begin{subfigure}[b]{0.15\textwidth}
        \includegraphics[width=\textwidth]{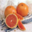}
        %\caption{CIFAR-10 VGG}
    \end{subfigure}
    \begin{subfigure}[b]{0.15\textwidth}
        \includegraphics[width=\textwidth]{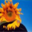}
        %\caption{CIFAR-10 VGG}
    \end{subfigure}
    \begin{subfigure}[b]{0.15\textwidth}
        \includegraphics[width=\textwidth]{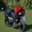}
        %\caption{CIFAR-10 VGG}
    \end{subfigure}
   \begin{subfigure}[b]{0.15\textwidth}
        \includegraphics[width=\textwidth]{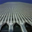}
        %\caption{CIFAR-10 VGG}
    \end{subfigure}
    \caption{Original Images}
    \label{fig:orig_img}
\end{figure*}

\begin{figure*}[b]
    \centering
    \begin{subfigure}[b]{0.15\textwidth}
        \includegraphics[width=\textwidth]{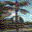}
        %\caption{CIFAR-10 VGG}
    \end{subfigure}
    \begin{subfigure}[b]{0.15\textwidth}
        \includegraphics[width=\textwidth]{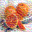}
        %\caption{CIFAR-10 VGG}
    \end{subfigure}
    \begin{subfigure}[b]{0.15\textwidth}
        \includegraphics[width=\textwidth]{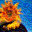}
        %\caption{CIFAR-10 VGG}
    \end{subfigure}
    \begin{subfigure}[b]{0.15\textwidth}
        \includegraphics[width=\textwidth]{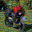}
        %\caption{CIFAR-10 VGG}
    \end{subfigure}
   \begin{subfigure}[b]{0.15\textwidth}
        \includegraphics[width=\textwidth]{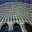}
        %\caption{CIFAR-10 VGG}
    \end{subfigure}
    \caption{Adversarial Images generated on SDNNs}
    \label{fig:et_img}
\end{figure*}

\begin{figure*}[t]
    \centering
    \begin{subfigure}[b]{0.15\textwidth}
        \includegraphics[width=\textwidth]{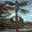}
        %\caption{CIFAR-10 VGG}
    \end{subfigure}
    \begin{subfigure}[b]{0.15\textwidth}
        \includegraphics[width=\textwidth]{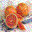}
        %\caption{CIFAR-10 VGG}
    \end{subfigure}
    \begin{subfigure}[b]{0.15\textwidth}
        \includegraphics[width=\textwidth]{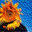}
        %\caption{CIFAR-10 VGG}
    \end{subfigure}
    \begin{subfigure}[b]{0.15\textwidth}
        \includegraphics[width=\textwidth]{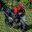}
        %\caption{CIFAR-10 VGG}
    \end{subfigure}
   \begin{subfigure}[b]{0.15\textwidth}
        \includegraphics[width=\textwidth]{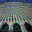}
        %\caption{CIFAR-10 VGG}
    \end{subfigure}
    \caption{Adversarial Images generated on DyNNs}
    \label{fig:ft_img}
\end{figure*}

In this section, we show original images, adversarial images generated through DyNNs and adversarial images generated through SDNNs through Figure \ref{fig:orig_img},Figure \ref{fig:ft_img} and Figure \ref{fig:et_img}. We find that in terms of quality, images generated through SDNNs (Average PSNR~\cite{fardo2016formal} = 23.20) are slightly better than images generated through DyNNs (Average PSNR = 23.19).

\section{RQ1 results based on Tiny Imagenet Images}

\begin{figure*}[t]
    \centering
    \begin{subfigure}[b]{0.465\textwidth}
        \includegraphics[width=\textwidth]{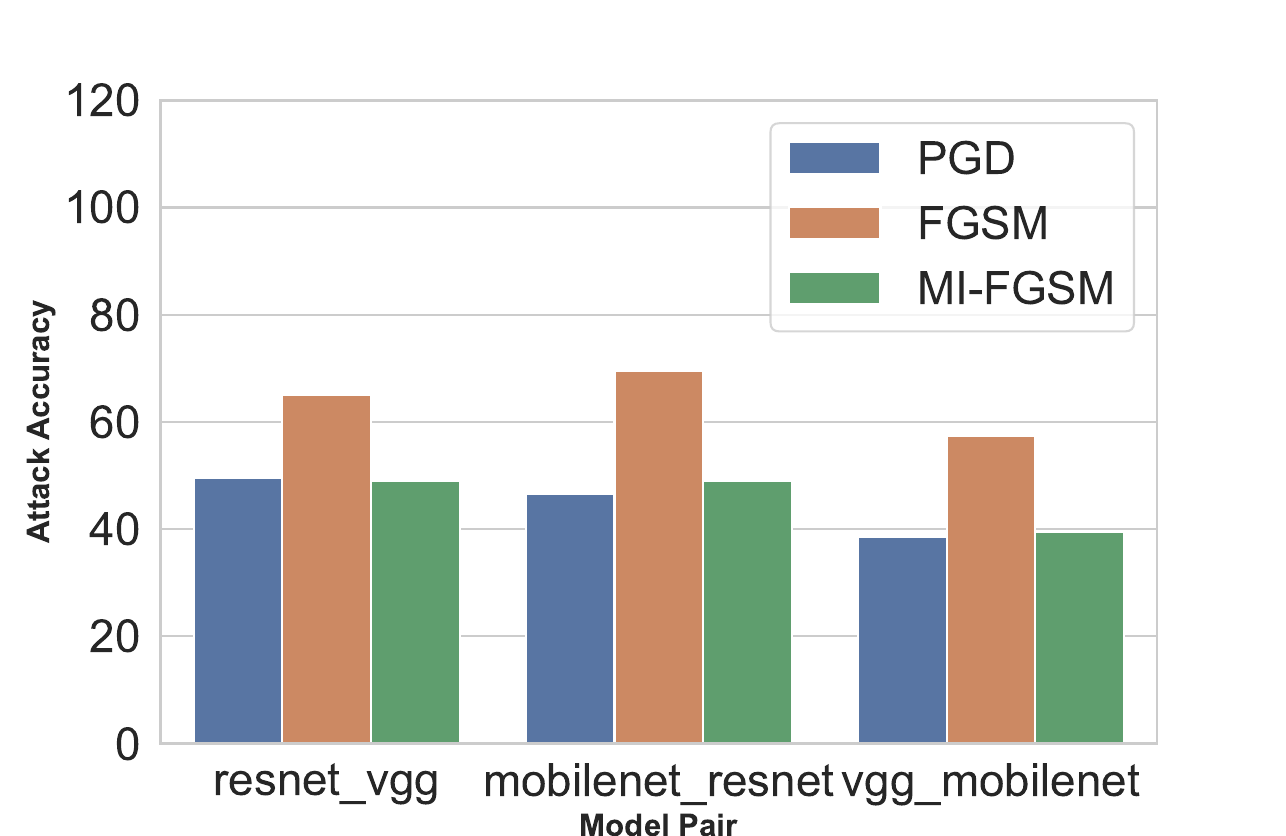}
        \caption{Target DyNN (Tiny Imagenet)}
    \end{subfigure}
    \begin{subfigure}[b]{0.45\textwidth}
        \includegraphics[width=\textwidth]{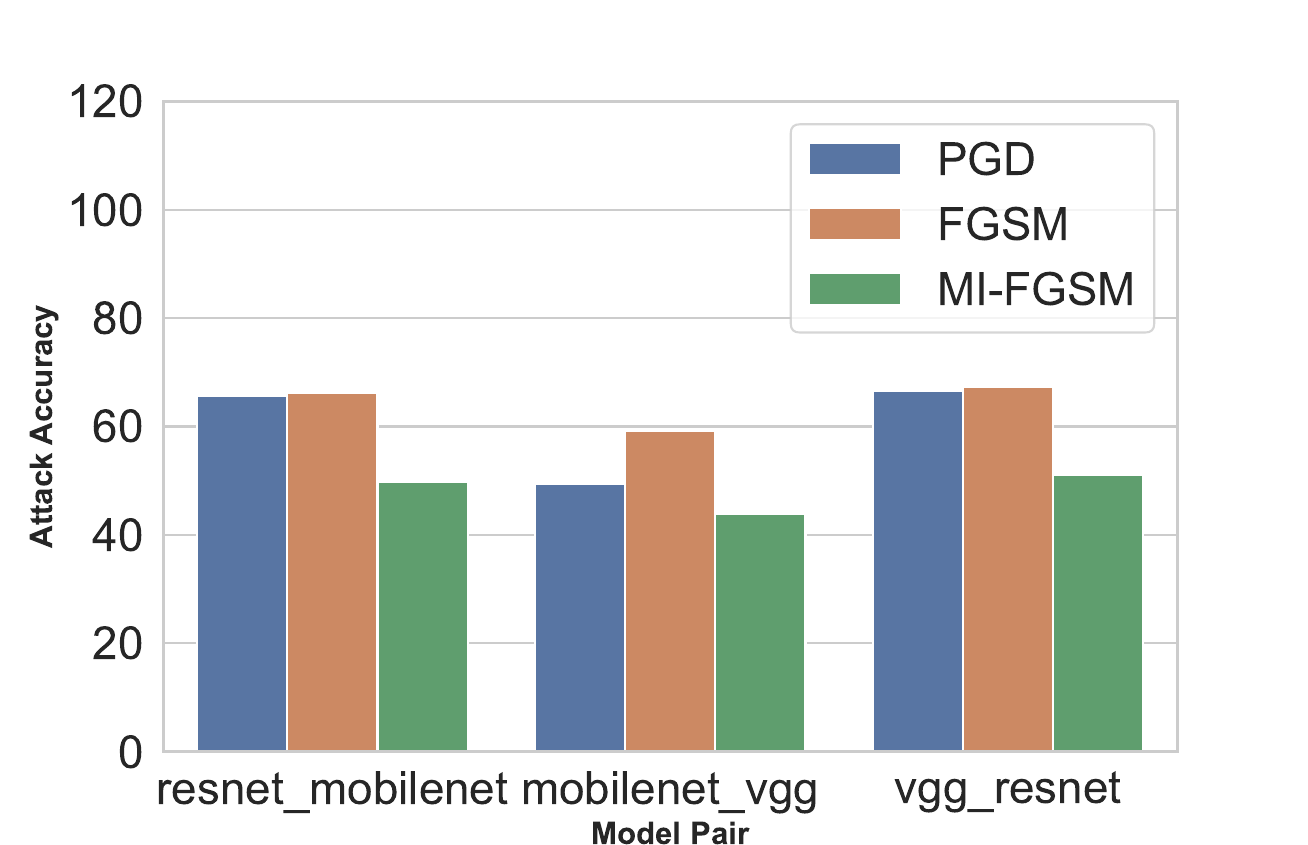}
        \caption{Target SDNN (Tiny Imagenet)}
    \end{subfigure}
    \caption{Transferable Attack Success Rate for Tiny Imagenet Data}
    \label{fig:tiny}
\end{figure*}

Here, through Figure \ref{fig:tiny}, we show the adversarial transferability results between DyNNs and SDNNs for Tiny Imagenet~\cite{chrabaszcz2017downsampled} datasets. These images are larger in size (64 $\times$ 64) than CIFAR images (size 32 $\times$ 32).
These results reconfirm that adversarial examples from DyNN to SDNN are more transferable than adversarial examples from SDNN to DyNN, even if the input feature space is larger. Although, we can notice a slight decrease in transferability for both D2S and S2D. 

%\appendix
%\section{Appendix}
%You may include other additional sections here.

\end{document}